\documentclass[A4paper]{article}

\usepackage{glossaries}
\usepackage{lscape}
\usepackage{rotating}
\usepackage{authblk}
\usepackage{amsmath,amssymb}
\usepackage{amsfonts} 
\usepackage{multirow}
\usepackage{booktabs}

\usepackage{xifthen}
\newcommand\lambdaone[1]{\ensuremath{\ifthenelse{\isempty{#1}}{}{（#1）}}}
\newcommand\lambdatwo[2]{\ensuremath{\ifthenelse{\isempty{#1#2}}{}{（#1, #2）}}}

\usepackage{tcolorbox}
    
\newcount\Comments  
\definecolor{darkgreen}{rgb}{0,0.5,0}
\definecolor{purple}{rgb}{1,0,1}
\newcommand{\kibitz}[2]{\ifnum\Comments=1\textcolor{#1}{#2}\fi}

\newcommand{\blue}[1]  {{#1}}

\usepackage[absolute,overlay]{textpos}
\usepackage{xpatch}
\usepackage{lscape}
\usepackage[misc,geometry]{ifsym}

\usepackage{multirow}

\usepackage{glossaries}
\newacronym{AE}{AE}{Adversarial Example}
\newacronym{OP}{OP}{Operational Profile}
\newacronym{NLP}{NLP}{Natural Language Processing}
\newacronym{RAM}{RAM}{Reliability Assessment Model}
\newacronym{ACU}{ACU}{Average Cell Unastuteness}
\newacronym{DL}{DL}{Deep Learning}
\newacronym{ML}{ML}{Machine Learning}
\newacronym{VAE}{VAE}{Variational Auto-Encoders}
\newacronym{VnV}{V\&V}{Verification and Validation}
\newacronym{KDE}{KDE}{Kernel Density Estimator}
\newacronym{PDF}{PDF}{Probability Density Function}
\newacronym{GA}{GA}{Genetic Algorithm}
\newacronym{FID}{FID}{Fréchet Inception Distance}
\newacronym{MSE}{MSE}{Mean Square Error}
\newacronym{OOD}{OOD}{Out-Of-Distribution}
\newacronym{HDA}{HDA}{Hierarchical Distribution-Aware}
\newacronym{GAN}{GAN}{Generative Adversarial Networks}
\newacronym{DNN}{DNN}{Deep Neural Network}
\newacronym{FGSM}{FGSM}{Fast Gradient Sign Method}
\newacronym{PGD}{PGD}{Projected Gradient Descent}

\newacronym{DRL}{DRL}{Deep Reinforcement Learning}
\newacronym{MDP}{MDP}{Markov Decision Process}
\newacronym{DDPG}{DDPG}{Deep Deterministic Policy Gradient}
\newacronym{PID}{PID}{Proportional-Integral-Derivative}
\newacronym{PMC}{PMC}{Probabilistic Model Checking}
\newacronym{DTMC}{DTMC}{Discrete Time Markov Chain}
\newacronym{CTMC}{CTMC}{Continuous Time Markov Chain}
\newacronym{LTL}{LTL}{Linear Temporal Logic}
\newacronym{PCTL}{PCTL}{Probabilistic Computational Tree Logic}
\newacronym{RAS}{RAS}{Robotics and Autonomous Systems}
\newacronym{vnv}{V\&V}{Verification and Validation}

\newacronym{MIQCP}{MIQCP}{Mixed Integer Quadratically Constrained Program}
\newacronym{IoGT}{IoGT}{Intersection-over-Ground-Truth}
\newacronym{IoU}{IoU}{Intersection-over-Union}
\newacronym{GIoU}{GIoU}{Generalized-IoU}
\newacronym{NN}{NN}{Neural Network}
\newacronym{AV}{AV}{Automated Vehicle}
\newacronym{NNs}{NNs}{Neural Networks}
\newacronym{AVs}{AVs}{Automated Vehicles}
\newacronym{AD}{AD}{Autonomous Driving}
\newacronym{ATN}{ATN}{Attention Network}
\newacronym{MLP}{MLP}{Multi-Layer Perceptron}
\newacronym{ATNs}{ATNs}{Attention Networks}
\newacronym{MLPs}{MLPs}{Multi-Layer Perceptrons}
\newacronym{MSA}{MSA}{Multi-Head Self-Attention}

\newacronym{LN}{LN}{Layer Normalization}
\newacronym{REG}{REG}{Regression}
\newacronym{CLS}{CLS}{Classification}
\newacronym{STN}{STN}{Spatial Transformation Network}
\newacronym{LDW}{LDW}{Lane Departure Warning}
\newacronym{GPU}{GPU}{Graphic Processing Unit}
\newacronym{GPUs}{GPUs}{Graphic Processing Units}
\newacronym{PO}{PO}{Potential Optimal}
\newacronym{CNN}{CNN}{Convolutional Neural Network}
\newacronym{RNNs}{RNNs}{Recurrent Neural Networks}
\newacronym{OODA}{OODA}{Out-Of-Distribution-Aware}
\newacronym{FODA}{FODA}{Feature-Only Distribution-Aware}
\newacronym{RQs}{RQs}{Research Questions}
\newacronym{PCA}{PCA}{Principal Component Analysis}
\newacronym{AVP}{AVP}{Autonomous Vehicle Parking}
\newacronym{PV}{PV}{Perspective-View}
\newacronym{BEV}{BEV}{Bird's-Eye-View}
\newacronym{BB}{BB}{Bounding Box}
\newacronym{mAP}{mAP}{mean Average Precision}
\newacronym{NMS}{NMS}{Non-Maximum Suppression}
\newacronym{SMT}{SMT}{Satisfiability Modulo Theories}

\usepackage{mathabx}
\usepackage{todonotes}
\usepackage{tikz}
\usepackage{hyperref}
\usepackage{tabularx}
\usepackage{subcaption} 
\usepackage{xpatch}
\usepackage{lscape}
\usepackage[misc,geometry]{ifsym}
\usepackage[absolute,overlay]{textpos}

\usepackage[T1]{fontenc}
\usepackage{graphicx}
\usepackage{amsmath}

\usepackage{url}
\usepackage{hyperref}
\hypersetup{urlcolor=blue}
\usepackage{graphicx}
\usepackage{graphics}
\usepackage{times}
\usepackage{mathptmx}
\usepackage{mathtools}
\usepackage[utf8]{inputenc}
\usepackage{caption}
\usepackage{subcaption}
\usepackage{booktabs}
\usepackage{multirow}
\usepackage{colortbl}
\usepackage{tikz}
\usepackage{pgfplots}
\usepackage{xfakebold}
\usepackage[makeroom]{cancel}

\newcommand{\setBoldness}[1]{\def\fake@bold{#1}}

\newcommand{\fbseries}{\unskip\setBold\aftergroup\unsetBold\aftergroup\ignorespaces}
\pgfplotsset{width=7.5cm,compat=1.12}
\usepgfplotslibrary{fillbetween}

\usepackage{algorithm}
\usepackage{algpseudocode}
\algnewcommand{\Initialize}[1]{
  \State \textbf{Initialize:}
  \Statex \hspace*{\algorithmicindent}\parbox[t]{.8\linewidth}{\raggedright #1}
}
\algnewcommand{\Input}[1]{
  \State \textbf{Input:} {\raggedright #1}
}
\algnewcommand{\Output}[1]{
  \State \textbf{Output:} {\raggedright #1}
}

\DeclareOldFontCommand{\rm}{\normalfont\rmfamily}{\mathrm}

\newacronym{llms}{LLMs}{Large Language Models}
\newacronym{llm}{LLM}{Large Language Model}

\begin{document}

\title{A Survey of 
Safety and Trustworthiness of 
Large Language Models through the Lens of Verification and Validation}


\author[1]{Xiaowei Huang}
\author[1]{Wenjie Ruan}
\author[5,1]{Wei Huang}
\author[1]{Gaojie Jin}
\author[1]{Yi Dong}
\author[2]{Changshun Wu}
\author[2]{Saddek Bensalem}
\author[1]{Ronghui Mu}
\author[1]{Yi Qi}
\author[1]{Xingyu Zhao}
\author[1]{Kaiwen Cai}
\author[1]{Yanghao Zhang}
\author[1]{Sihao Wu}
\author[1]{Peipei Xu}
\author[1]{Dengyu Wu}
\author[3]{Andre Freitas}
\author[3,4]{Mustafa A. Mustafa}

\affil[1]{University of Liverpool, UK}
\affil[2]{Université Grenoble Alpes, France}
\affil[3]{The University of Manchester, UK}
\affil[4]{COSIC, KU Leuven, Belgium}
\affil[5]{Purple Mountain Laboratories, China}

\date{}

\maketitle

\begin{abstract}
\gls{llms} have exploded a new heatwave of AI for their ability to engage end-users in human-level conversations  with 
detailed  and articulate answers across many 
knowledge domains. In response to their fast adoption in many industrial applications, this survey concerns their safety and trustworthiness. 
First, we review known vulnerabilities and limitations of the \gls{llms}, categorising them into inherent issues, attacks, and unintended bugs. Then, we consider if and how the \gls{vnv} techniques, which have been widely developed for traditional software and deep learning models such as convolutional neural networks \blue{as independent processes to check the alignment of their implementations against the specifications}, can be integrated and further extended throughout the lifecycle of the \gls{llms} to provide rigorous analysis to the safety and trustworthiness of \gls{llms} and their applications. Specifically, we consider four complementary techniques: falsification and evaluation, verification, runtime monitoring, and regulations and ethical use. In total, 370+ references are considered to support the quick understanding of the safety and trustworthiness issues from the perspective of \gls{vnv}.
\blue{While intensive research has been conducted to identify the safety and trustworthiness issues, rigorous yet practical methods are called for to ensure the alignment of \gls{llms} with  safety and trustworthiness requirements.} 

\end{abstract}

\tableofcontents


\section{Introduction}

A \gls{llm} is a deep learning model equipped with a massive amount of learnable parameters (commonly reaching more than 10 billion).
LLMs are attention-based sequential models based on the transformer architecture~\cite{hrinchuk2020correction}, which consistently demonstrated the ability to learn universal representations of language. The universal representations of language  can then be used in various \gls{NLP} task. 
The recent scale-up of these models, in terms of both numbers of parameters and pre-trained corpora, has 
confirmed the universality of transformers as mechanisms to encode language representations. At a specific scale, these models started to exhibit in-context learning~\cite{min2022rethinking,ye2022complementary}, and the properties of learning from few examples (zero/one/few-shot -- without the need for fine-tuning) and from natural language prompts (complex instructions which describe the behavioural intent that the model needs to operate). Recent works on Reinforcement Learning via Human Feedback (RLHF)~\cite{ouyang2022training} have further developed the ability of these models to align and respond to increasingly complex prompts, leading to their popularisation in systems such as ChatGPT~\cite{ChatGPT} and their use in a large spectrum of applications.
The ability of LLMs to deliver sophisticated linguistic and reasoning behaviour, has pushed their application beyond their intended operational envelope. 

While being consistently fluent, LLMs are prone to hallucinations 
\cite{shuster2021retrieval}, stating factually incorrect statements 
\cite{shuster2022language}, lacking necessary mechanisms of safety, lacking transparency and control 
\cite{tanguy2016natural}, among many others. \blue{Such vulnerabilities and limitations have already led to bad consequences such as suicide case \cite{vicekill}, lawyer submitted fabricated cases as precedent to the court \cite{lawyercasebbc}, leakage of private information \cite{ChatGPTchathistorybug}, etc.   Therefore, research is urgently needed to understand the potential vulnerabilities and how the \gls{llm}s' behaviour can be assured to be safe and trustable.}
The goal of this paper is to provide a review of known vulnerabilities and limitations of \gls{llm}s and, more importantly, to investigate how the \gls{vnv} techniques can be adapted to improve the safety and trustworthiness of \gls{llm}s. While there are several surveys on \gls{llm}s \cite{zhou2023comprehensive,zhao2023survey}, as well as a categorical archive of ChatGPT failures \cite{borji2023categorical}, to the best of our knowledge, this is the first work that provides a comprehensive discussion on the safety and trustworthiness issues, from the perspective of the \gls{vnv}. 

\begin{figure}[htp]
    \centering
    \includegraphics[width=1\textwidth]{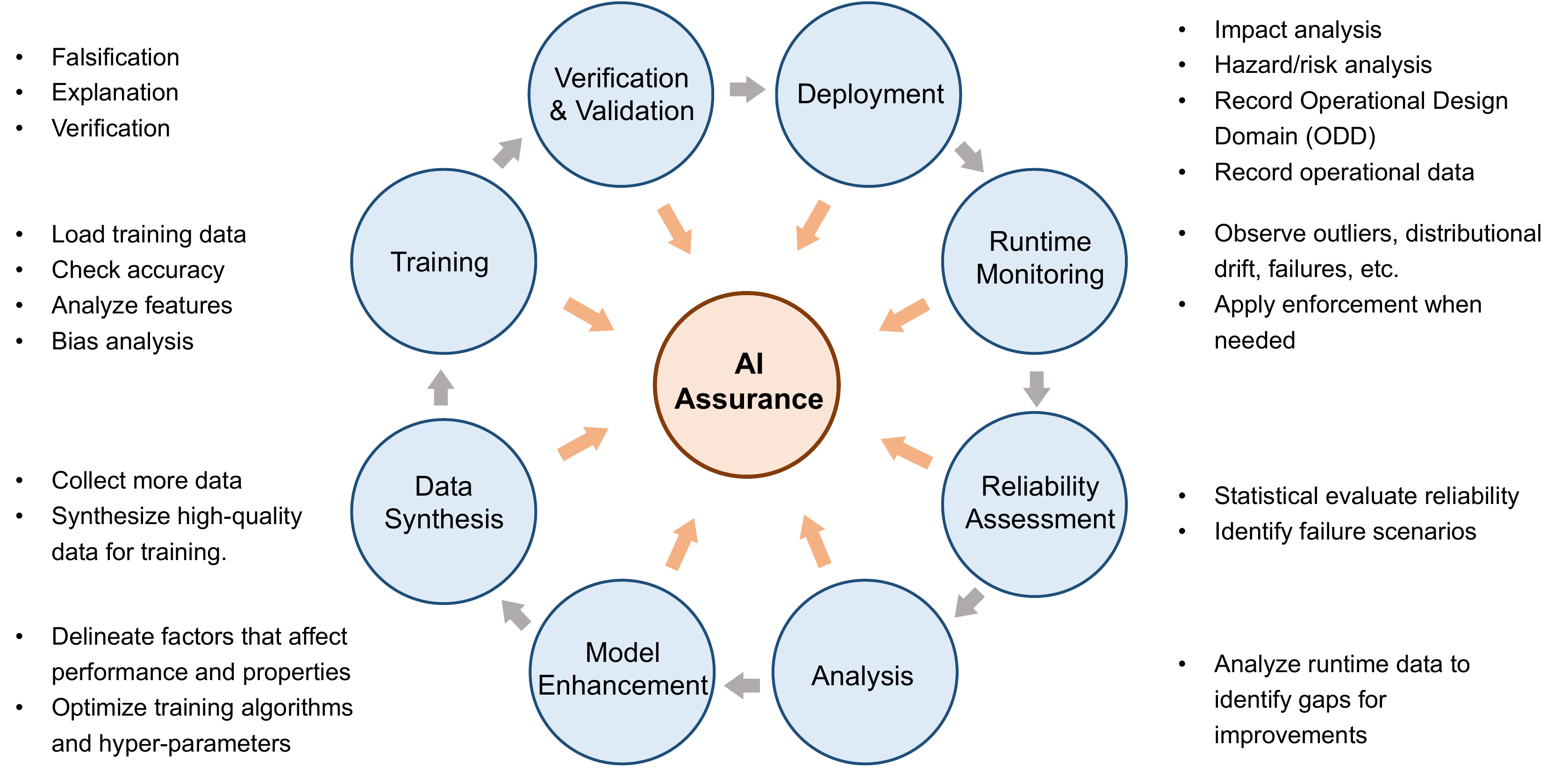}
    \caption{Summarisation of lifecycle V\&V methods to support AI Assurance.}
    \label{fig:lifecycle}
\end{figure}

With the rising of \gls{llm}s and its wide applications, the need to ensure their safety and trustworthiness become prominent. 
Considering the broader subject of deep learning systems, to support their safety and trustworthiness, a diverse set of technical solutions have been developed by different research communities. For example, the machine learning community is focused on adversarial attacks \cite{goodfellow2014explaining, madry2017towards, croce2020reliable, xu2020adversarial},  outlier detectors \cite{pang2021deep}, adversarial training \cite{szegedy2013intriguing, pmlr-v80-mirman18b, wong2020fast}, and explainable AI \cite{xu2019explainable, gunning2019xai, lime, pmlr-v161-zhao21a}. The human-computer interaction community is focused on engaging the learning systems in the interactions with end users to improve the end users' confidence \cite{dudley2018review}. Formal methods community treats ML models as yet another symbolic system (evidenced by their consideration of neurons, layers, etc.) and adapts existing formal methods tools to work on the new systems \cite{HUANG2020100270}. While research has been intense on these individual methods, a synergy among them has not been addressed. \emph{Without a synergy, it is hard, if not impossible, to rigorously understand the causality between methods and how they might collectively support the safe and trusted autonomous systems at runtime}. This survey is rooted in the field of AI assurance, aiming to apply a collection of rigorous \gls{vnv} methods throughout the lifecycle of ML models, to provide assurance on the safety and trustworthiness. An illustrative diagram is given in Figure~\ref{fig:lifecycle} for general ML models. \blue{To begin with, data collection and synthesis is required to obtain as many as possible the training data, including the synthesis of high quality data through e.g., data argumentation or generative models. 
In the training phase, other than the prediction accuracy, multiple activities are needed, including e.g., the analysis of the learned feature representations and the checking for unintended bias. 
After the training, we apply offline \gls{vnv} methods to the ML model, including techniques to falsify, explain, and verify the ML models. During the deployment phase, we must analyse the impact and hazard of the potential application environment. The operational design domain and operational data will be recorded. A run-time monitor is associated to the ML model to detect outliers, distribution shifts, and failures in the application environment. We may further apply reliability assessment methods to evaluate the reliability of the ML model and identify failure scenarios. Based on the detection or assessment results, we can identify the gaps for improvement. Finally, we outline the factors that affect the performance of the ML model, and optimise the training algorithm to obtain an enhanced ML model.} 



These \gls{vnv} techniques have been successful in supporting the reliable and dependable development of software and hardware that are applied to safety-critical systems, and have been adapted to work with machine learning models, mainly focusing on the convolutional neural networks for image classification (see surveys such as \cite{HUANG2020100270,liu2020algorithms} and textbooks such as \cite{MLSafety2023}), but also extended to consider, for example, object detection, deep reinforcement learning, and recurrent neural networks. This paper discusses how to extend further the \gls{vnv} techniques to deal with the safety and trustworthiness challenges of \gls{llm}s. 

\gls{vnv} are independent procedures that are used together for checking that a system (or product, service) meets requirements and specifications and that it fulfills its intended purpose \cite{vnv2004}. Among them, verification techniques check the system against a set of design specifications, and validation techniques ensure that the system meets the user's operational needs. From software, convolutional neural networks to \gls{llm}s, the scale of the systems grows significantly, which makes the usual \gls{vnv} techniques less capable due to their computational scalability issues. White-box \gls{vnv} techniques that take the learnable parameters as their algorithmic input will not work well in practice. Instead,  the research should focus on black-box techniques, on which some research has started for convolutional neural networks. In addition, \gls{vnv} techniques need to consider the \emph{non-deterministic nature} of \gls{llm}s (i.e., different outputs for two tests with identical input), which is a noticeable difference with the usual neural networks, such as convolutional neural networks and object detectors, that currently most \gls{vnv} techniques work on.

\blue{Considering the fast development of \gls{llms}, this survey does not intend to be complete (although it includes 370+ references), especially when it comes to the applications of \gls{llms} in various domains, but rather a collection of organised literature reviews and discussions to support the 
understanding of the safety and trustworthiness issues from the perspective of \gls{vnv}. Through the survey, we noticed that the current research are  focused on identifying the vulnerabilities, with limited efforts on systematic approaches to evaluate and verify the safety and trustworthiness properties. }

The structure of the paper is as follows. In Section~\ref{sec:llms}, we review the \gls{llm}s and its categories, its lifecycle, and several techniques introduced to improve safety and trustworthiness. Then, in Section~\ref{sec:vulnerabilities}, we present a review of existing vulnerabilities. This is followed by a general verification framework in Section~\ref{sec:generalframework}. The framework includes \gls{vnv} techniques such as falsification and evaluation (Section~\ref{sec:falsification}), verification (Section~\ref{sec:verification}), runtime monitor (Section~\ref{sec:runtime}), and ethical use (Section~\ref{sec:ethical}). We conclude the paper in Section~\ref{sec:concl}.

\section{Large Language Models}\label{sec:llms}

This section summarises the categories of machine learning tasks based on \gls{llm}s, followed by a discussion of the lifecycle of \gls{llm}s. We will also discuss a few fundamental techniques relevant to the safety analysis. 

\subsection{Categories of Large Language Models}

\gls{llm}s have been applied to many tasks, such as text generation \cite{li2022pretrained}, content summary \cite{zhang2023benchmarking}, conversational AI (i.e., chatbots) \cite{wei2023leveraging}, and image synthesis \cite{koh2023generating}. Other \gls{llm}s applications can be seen as their adaptations or further applications. In the following, we discuss the two most notable categories of \gls{llm}s: \blue{text-based conversational AI and image synthesis}. \blue{While they might have slightly different concerns, this survey will be more focused on issues related to the former, without touching some issues that are specific to image synthesis such as the detection of fake images. } 


\subsubsection{Text-based Conversational AI}

LLMs are 
designed to understand natural language and generate human-like responses to queries and prompts. 
Almost all \gls{NLP} tasks (e.g., language translation \cite{brants2007large}, chatbots \cite{lin2019caire,10.1145/3340531.3412330} and virtual assistants \cite{tulshan2019survey}) have witnessed tremendous success with Transformer-based pretrained language models (T-PTLMs), relying on Transformer~\cite{NIPS2017_3f5ee243}, self-supervised learning~\cite{jaiswal2020survey,liu2021self} and transfer learning~\cite{houlsby2019parameter,ruder2019transfer} to process and understand the nuances of human language, including grammar, syntax, and context.

The well-known text-based \gls{llm}s include GPT-1 \cite{radford2018improving}, BERT \cite{devlin-etal-2019-bert}, XLNet \cite{yang2019xlnet}, RoBERTa \cite{liu2019roberta}, ELECTRA \cite{clark2020electra}, T5 \cite{raffel2020exploring}, ALBERT \cite{lanalbert}, BART \cite{lewis-etal-2020-bart}, and PEGASUS \cite{pmlr-v119-zhang20ae}. These models can learn general language representations from large volumes of unlabelled text data through self-supervised learning and subsequently transfer this knowledge to specific tasks, which has been a major factor contributing to their success in NLP \cite{kalyan2021ammus}. Kaplan et al. \cite{kaplan2020scaling} demonstrated that simply increasing the size of T-PTLMs can lead to improved performance \cite{kalyan2021ammus}. This finding has spurred the development of LLMs such as GPT-3 \cite{NEURIPS2020_1457c0d6}, PANGU \cite{https://doi.org/10.48550/arxiv.2104.12369}, GShard \cite{lepikhin2020gshard}, Switch-Transformers \cite{fedus2021switch} and GPT-4 \cite{openai2023gpt4}. 

\begin{figure*}
    \centering
    \includegraphics[width=0.8\textwidth]{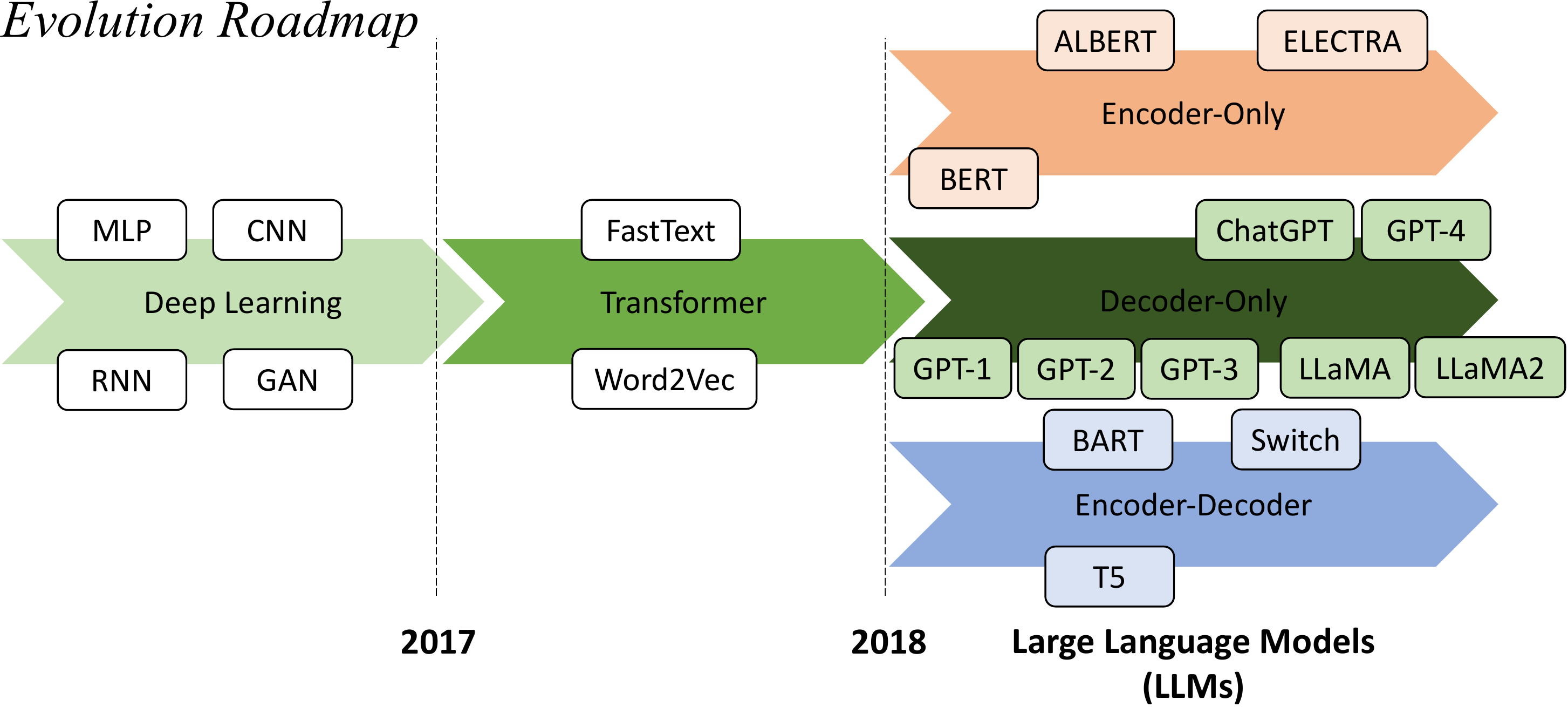}
    \caption{Large Language Models: Evolution Roadmap.}
    \label{fig:llm_evolution}
\end{figure*}

\blue{With the advancement of the Transformer development \cite{NIPS2017_3f5ee243}, significant enhancements were achieved in handling sequential data. Leveraging the Transformer architecture, \gls{llm}s have been created as potent models with the capacity to generate text resembling human language. ChatGPT represents a distinct embodiment of an LLM, characterised by its remarkable performance that yields groundbreaking outcomes. The progression of \gls{llm}s, depicted in Figure~\ref{fig:llm_evolution}, starts from the evolution of deep learning and transformer-based frameworks, culminating in the latest explosion of \gls{llm}s. We divide the \gls{llm}s into Encoder-only, Decoder-only, and Encoder-Decoder according to \cite{yang2023harnessing}. In Encoder-only and Encoder-Decoder architectures, the model predicts masked words in a sentence while taking into account the surrounding context. While Decoder-only models are trained by generating the subsequent word in a sequence based on the preceding words. GPT-style language model belongs to the Decoder-only type.}  

\blue{We also note that, there are advanced uses of \gls{llms} (or advanced prompt engineering) by considering e.g., self-consistency \cite{wang2023selfconsistency}, knowledge graph \cite{pan2023unifying}, generating programs as the intermediate reasoning steps \cite{gao2023pal}, generating both reasoning traces and task-specific actions in an interleaved manner \cite{yao2023react}, etc.}

\subsubsection{Text-based Image Synthesis} 

The transformer model \cite{vaswani2017attention} has become the standard choice for Language Modelling tasks, but it has also found widespread integration in text-to-image tasks. We present a chronological overview of the advancements in text-to-image research. 
DALL-E \cite{ramesh2021zero} is a representative approach that leverages Transformers for a text-to-image generation. The methodology involves training a dVAE \cite{rolfe2016discrete} and subsequently training a 12B decoder-only sparse transformer  supervised by image tokens from the pre-trained dVAE. The transformer generates image tokens solely based on text tokens during inference. The resulting image candidates are evaluated by a pretrained CLIP model~\cite{radford2017learning} to produce the final generated image.
StableFusion~\cite{rombach2022high} differs from DALL-E~\cite{ramesh2021zero} by using a diffusion model instead of a Transformer to generate latent image tokens. To incorporate text input, StableFusion~\cite{rombach2022high} first encodes the text using a transformer then conditions the diffusion model on the resulting text tokens.
GLIDE~\cite{nichol2021glide} employs a transformer model \cite{vaswani2017attention} to encode the text input and then trains a diffusion model to generate images that are conditioned on the text tokens directly.
DALL-E2~\cite{ramesh2022hierarchical} effectively leverages LLMs by following a three-step process. First, a CLIP model is trained using text-image pairs. Next, using text tokens as input, an autoregressive or diffusion model generates image tokens. Finally, based on these image tokens, a diffusion model is trained to produce the final image. 
Imagen~\cite{saharia2022photorealistic} employs a pre-trained text encoder, such as BERT~\cite{devlin2018bert} or CLIP \cite{radford2017learning}, to encode text. It then uses multiple diffusion models to train a process that generates images that start from low-resolution and gradually progress to high-resolution.
Parti~\cite{yu2022scaling} demonstrates that a VQGAN~\cite{esser2021taming} and Transformer architecture can achieve superior image synthesis outcomes compared to previous approaches, even without utilising a diffusion model. 
The eDiff-I model~\cite{balaji2022ediffi} has recently achieved state-of-the-art performance on the MS-COCO dataset \cite{lin2014microsoft} by leveraging a combination of CLIP and diffusion models. 

In summary, text-to-image research commonly utilises transformer models \cite{vaswani2017attention} for encoding text input and either the diffusion model or the decoder of an autoencoder for generating images from latent text or image tokens.

\subsection{Lifecycle of \gls{llms}}



\begin{figure*}
    \centering
    \includegraphics[width=\textwidth]{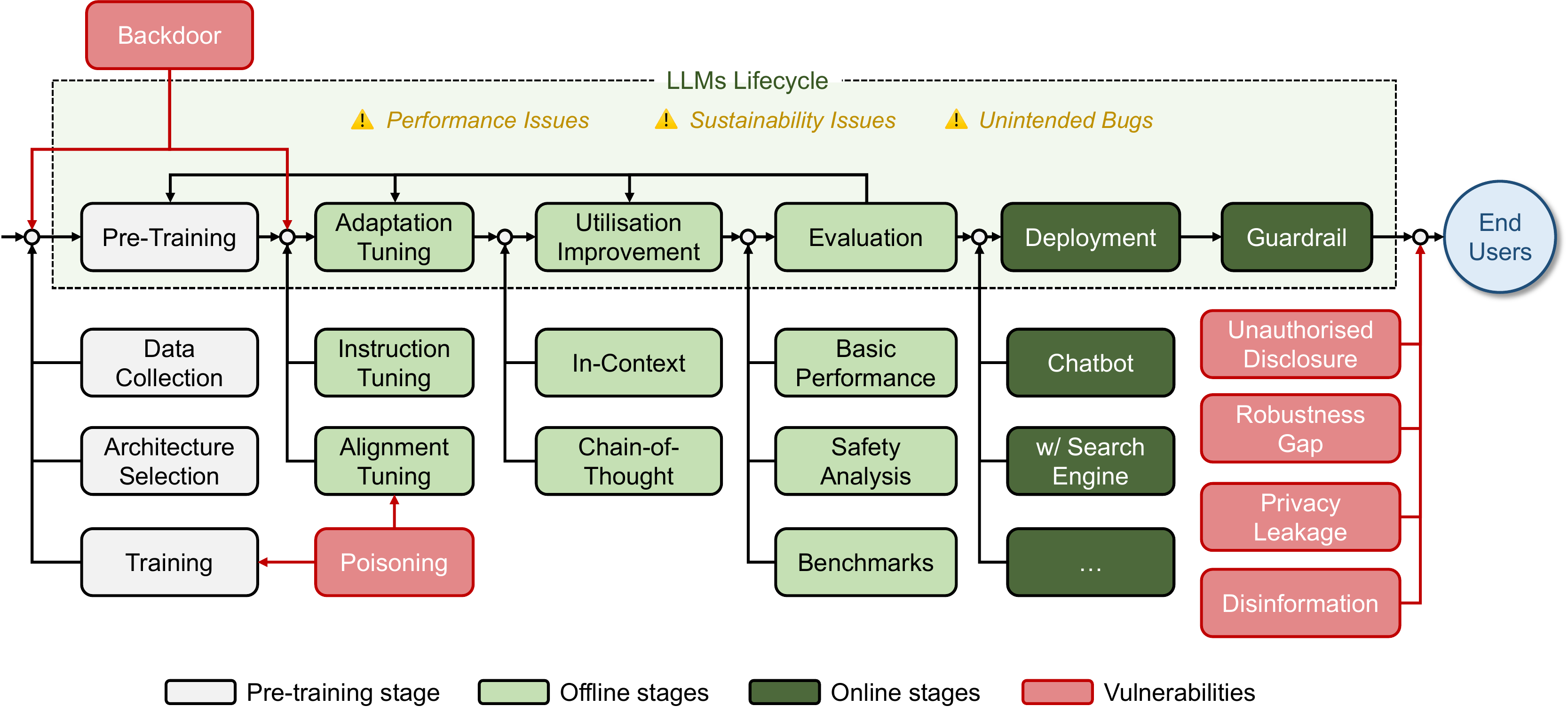}
    \caption{Large Language Models: Lifecycle and Vulnerabilities.}
    \label{fig:framework}
\end{figure*}

Figure~\ref{fig:framework} illustrates the lifecycle stages and the vulnerabilities of \gls{llm}s. \blue{This section will focus on the introduction of lifecycle stages, and the detailed discussions about vulnerabilities will appear in Section~\ref{sec:vulnerabilities}.}
The offline model construction is formed of three steps \cite{zhao2023survey}: pre-training, adaptation tuning, and utilisation improvement, such that each step includes several interleaving sub-steps. In general, the \emph{pre-training} step is similar to the usual machine learning training that goes through data collection, architecture selection, and training. On \emph{adaptation tuning}, it might conduct instruction tuning \cite{lou2023prompt} to learn from task instructions, and alignment tuning \cite{ouyang2022training,NIPS2017_d5e2c0ad} to make sure  \gls{llm}s are aligned with human values, e.g., fair, honest, and harmless. Beyond this, to improve the interaction with the end users, \emph{utilisation improvements} may be conducted through, for example, in-context learning \cite{NEURIPS2020_1457c0d6} and  chain-of-thought learning \cite{wei2022chain}.  

Once an \gls{llm} is trained,
an \emph{evaluation} is needed to ensure that its performance matches the expectation. Usually, we consider the evaluation from three perspectives: evaluation on basic performance, safety analysis to evaluate the consequence of applying the LLM in an application, and the evaluation through publicly available benchmark datasets. 
\blue{The basic performance evaluation  considers several basic types of abilities such as language generation and complex reasoning. Safety analysis is to understand the impacts of human alignment, interaction with external environment, and incorporation of \gls{llm}s into broader applications such as search engines. On top of these, benchmark datasets and publicly available tools are used as well to support the evaluation. } 
The evaluation will determine if the LLM is acceptable (for pre-specified criteria), and if so, the process will move forward to the deployment stage. Otherwise, at least one failure will be identified, and the process will move back to either of the three training steps. 

On the \emph{deployment} stage, we will determine how the LLM will be used. For example, it could be available in a web platform for direct interaction with end users, such as the ChatGPT\footnote{\url{https://openai.com/blog/ChatGPT}}. Alternatively, it may be embedded into a search engine, such as the new Bing\footnote{\url{https://www.bing.com/new}}. Nevertheless, according to the common practice, a \emph{guardrail} is imposed on the conversations between \gls{llm}s and end users to ensure that the AI regulation is maximally implemented. 


\blue{In Figure 2, within the LLMs lifecycle, three main issues run through: performance issues, sustainability issues, and unintended bugs. These may be caused by one or more stages in the lifecycle. The red block shows that vulnerabilities appear in the LLMs lifecycle, and they may appear in the early stage of the whole period. For example, backdoor attacks and poisoning can contaminate raw data. When LLMs are deployed, problems such as a robustness gap may also arise.}

\subsection{Key Techniques Relevant to Safety and Trustworthiness}

In the following, we discuss two fundamental techniques that are distinct from the usual deep learning models and have been used by e.g., ChatGPT to improve safety and trustworthiness: \blue{reinforcement learning from human feedback and guardrails}. 

\subsubsection{Reinforcement learning from human feedback (RLHF)}


\blue{RLHF can be conducted in any stage of the ``Adapation Tuning'', ``Utilisation Improvement'', or ``Evaluation'' in the framework of Figure~\ref{fig:framework}}. RLHF \cite{christiano2017deep, ouyang2022training, hh_assistant, bai2022training, bai2022constitutional, openai2023gpt4, lambert2022illustrating, ziegler2019fine} plays a crucial role in the training of language models, as it allows the model to learn from human guidance and avoid generating harmful content. In essence, RLHF assists in aligning language models with safety considerations through  fine-tuning with human feedback. OpenAI initially introduced the concept of incorporating human feedback to tackle complex reinforcement learning tasks in \cite{christiano2017deep}, which subsequently facilitated the development of more sophisticated \gls{llms}, from InstructGPT \cite{ouyang2022training} to GPT4 \cite{openai2023gpt4}. According to InstructGPT \cite{ouyang2022training}, the RLHF  training process typically begins by learning a reward function intended to reflect what humans value in the task, utilising human feedback on the model's outputs. Subsequently, the language model is optimised via an RL algorithm, such as PPO~\cite{schulman2017proximal}, using the learned reward function. Reward model training and fine-tuning with RL can be iterated continuously. More comparison data is collected on the current best policy, which is used to train a new reward model and  policy.
The InstructGPT models demonstrated enhancements in truthfulness and reductions in generating toxic outputs  while maintaining minimal performance regressions on public NLP datasets. 

Following InstructGPT, Red Teaming language models \cite{bai2022training} introduces a harmlessness preference model to help RLHF to get less harmful agents. The comparison data from red team attacks is used as the training data to develop the harmlessness preference model. \blue{The authors of}~\cite{hh_assistant} utilised the helpful and harmless datasets in preference modelling and RLHF to fine-tune \gls{llms}. They discovered that there was a significant tension between helpfulness and harmlessness. Experiments showed helpfulness and harmlessness model is significantly more harmless than the model trained only on helpfulness data. 
They also found that alignment with RLHF has many benefits and no cost to performance, like combining alignment training with programming ability and summarisation. 
\blue{The authors of}~\cite{ganguli2023capacity} found that \gls{llms} trained with RLHF have the capability for moral self-correction. They believe that the models can learn intricate normative concepts such as stereotyping, bias, and discrimination that pertain to harm. Constitutional AI \cite{bai2022constitutional} trains the preference model by relying solely on AI feedback, without requiring  human labels to identify harmful outputs. To push the process of aligning \gls{llms} with RLHF, an open-sourced modular library, RL4LMs, and evaluation benchmark, GRUE, designed for optimising language generator with RL are introduced in \cite{ramamurthy2022reinforcement}. Inspired by the success of RLHF in language-related domains, fine-tuning approaches that utilise human feedback to improve text-to-image models \cite{lee2023aligning, xu2023imagereward, wu2023better} have gained popularity as well. To achieve human-robot coexistence, \blue{the authors of}~\cite{gu2023human} proposed a human-centred robot RL framework consisting of safe exploration, safety value alignment, and safe collaboration. They discussed the importance of interactive behaviours and four potential challenges within human-robot interactive procedures. Although many works indicate that RLHF could decrease the toxicity of generations from \gls{llms}, the induced RLHF, like introducing malicious examples by annotators \cite{carlini2023poisoning}, may cause catastrophic performance and risks. We hope better techniques that lead to transparency, safe and trustworthy RLHF will be developed in the coming future.


\subsubsection{Guardrails}\label{sec:guardrail}

Considering that some \gls{llms} are interacting directly with end-users, it is necessary to put a layer of protection\blue{, called guardrail,} when the end users ask for information about violence, profanity, criminal behaviours, race, or other unsavoury topics. \blue{Guardrails are deployed in most, if not all, \gls{llms}, including ChatGPT, Claude, and LLaMA.} In such cases, a response is provided with the \gls{llm} refusing to provide information. While this is a very thin layer of protection because there are many tricks (such as prompt injections that will be reviewed in Section~\ref{sec:promptInjection}) to circumvent it, it enhances the social responsibility of \gls{llm}s. 

\section{Vulnerabilities\blue{, Attacks, and Limitations}}\label{sec:vulnerabilities}

\begin{figure}
    \centering
    \includegraphics[width=0.8\textwidth]{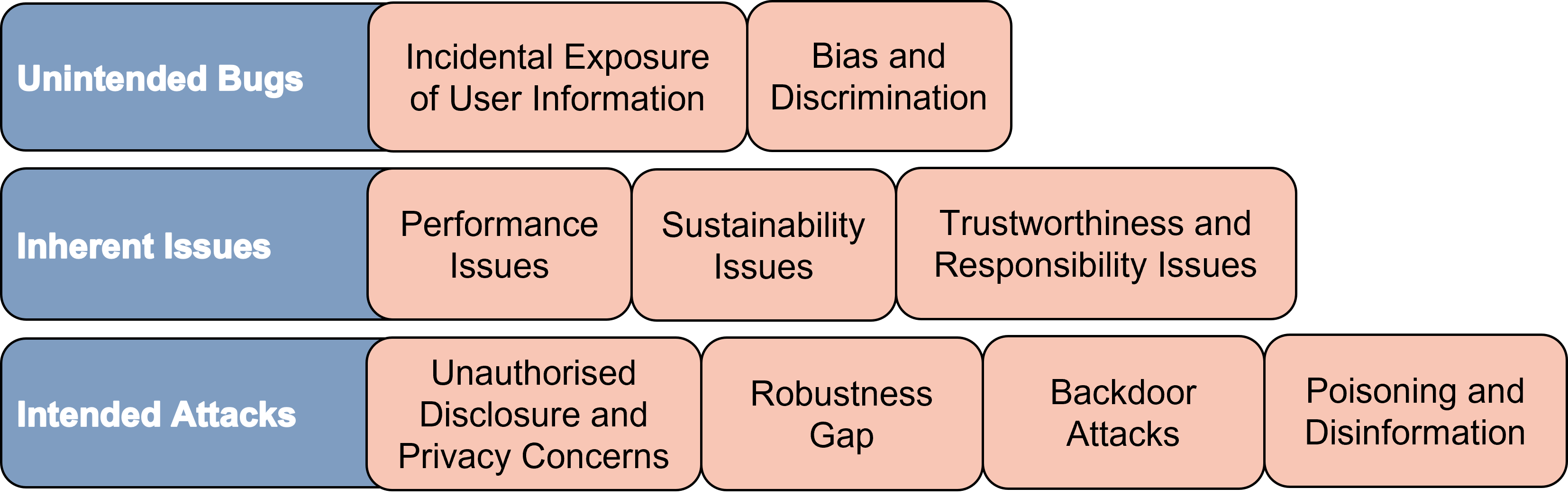}
    \caption{Taxonomy of Vulnerabilities.}
    \label{fig:taxonomy}
\end{figure}


This section presents a review of the known types of vulnerabilities. The vulnerabilities can be categorised into inherent issues, intended attacks, and unintended bugs, as illustrated in Figure~\ref{fig:taxonomy}. 

\emph{Inherent issues} are vulnerabilities that cannot be readily solved by the \gls{llm}s themselves. However, they can be gradually improved with, e.g., more data and novel training methods. Inherent issues include performance weaknesses, which are those aspects that \gls{llm}s have not reached the human-level intelligence, and sustainability issues, which are because the size of \gls{llm}s is significantly larger than the usual machine learning models. Their training and daily execution can have non-negligible sustainability implications. Moreover, trustworthiness and responsibility issues are inherent to the \gls{llm}s. 

\emph{Attacks} are initiated by malicious attackers, which attempt to implement their goals by attacking certain stages in the \gls{llm}s lifecycle. Known intended attacks include robustness gap, backdoor attack, poisoning, disinformation, privacy leakage, and unauthorised disclosure of information. 

Finally, with the integration of \gls{llm}s into broader applications, there will be more and more \emph{unintended bugs} that are made by the developers unconsciously but have serious consequences, such as bias and discrimination (that are usually related to the quality of training data), and the recently reported incidental exposure of user information. \blue{We separate these from inherent issues, because they could be resolved with e.g., high quality training data, carefully designed API, and so on. They are ``unintended'', because they are not deliberately designed by the developers. }

Figure~\ref{fig:framework} suggests how the vulnerabilities may be exploited in the lifecycle of \gls{llm}s. While inherent issues and unintended bugs may appear in any stage of the lifecycle, the  attacks usually appear in particular stages of the lifecycle. For example, a backdoor attack usually occurs in pre-training and adaptation tuning, in which the backdoor trigger is embedded, and poisoning usually happens in training or alignment tuning, when the \gls{llm}s acquires information/data from the environment. Besides, many attacks occur upon the interaction between end users and the \gls{llm}s using specific, well-designed prompts to retrieve information from the \gls{llm}s. We remark that, while there are overlapping, \gls{llm}s and usual deep learning models (such as convolutional neural networks or object detectors) have slightly different vulnerabilities, and while initiatives have been taken on developing specification languages for usual deep learning models \cite{BCHKMNP2022,10.1007/978-3-031-17244-1_1}, such efforts may need to be extended to \gls{llm}s. 


\subsection{Inherent Issues}

\subsubsection{Performance Issues}\label{sec:performanceweaknesses}

Unlike traditional software systems, which run according to the rules that can be deterministically verified, neural network-based deep learning systems, including large-scale \gls{llm}s, have their behaviours determined by the complex models learned from data through optimisation algorithms. It is unlikely that an \gls{llm} performs 100\% correctly.
\blue{As a simple example shown in Table \ref{error_example}, it can be observed that similar errors exist across different \gls{llm}s, where most of the existing \gls{llm}s are not able to provide a correct answer.}
Performance issues \blue{related to the correctness of the outputs} include at least the following two categories: factual errors and reasoning errors.

\begin{table*}[!h]
\centering
\caption{\blue{Performance error exists across different \gls{llm}s. Retrieved 24 August 2023.}}
\label{error_example}
\resizebox{1\linewidth}{!}{
\renewcommand\arraystretch{1.5}
\begin{tabular}{l|c}
\toprule[1.5pt]
\blue{\gls{llm}s}      &\blue{Output for question:  "Adam’s wife is Eve. Adam’s daughter is Alice. Who is Alice to Eve?"}
\\
\midrule[1.5pt]
\blue{ChatGPT \cite{ChatGPT}}   &  \blue{Alice is Eve's granddaughter.} \\ \hline
\blue{ERNIE Bot \cite{Yang_2023}} &  \blue{Alice is Eve's granddaughter.}   \\ \hline
\blue{Llama2 \cite{touvron2023llama}} &  \blue{Alice is Eve's granddaughter.}    \\ \hline
\blue{Bing Chat \cite{Mehdi_2023}} &  \blue{Alice is Adam’s daughter and Eve’s granddaughter.}\\\hline 
\blue{GPT-4 \cite{openai2023gpt4}} &\blue{Alice is Eve's daughter.}\\
\bottomrule[1.5pt]
\end{tabular}
}
\end{table*}

\paragraph{Factual errors} Factual errors refer to situations where the output of an \gls{llm} contradicts the truth, \blue{where some literature refers this situation as \textit{hallucination} \cite{openai2023gpt4}\cite{zhao2023survey}\cite{li2023halueval}}. For example, when asked to provide information about the expertise in the computer science department at the University of Liverpool, the ChatGPT refers to people who were never affiliated with the department. 
Hence more serious errors can be generated, including notably wrong medical advice. Additionally, it is interesting to note that while \gls{llm}s can perform across different domains, their reliability may vary across domains. For example, \blue{the authors of}~\cite{shen2023ChatGPT} show that ChatGPT significantly under-performs in law and science questions. Investigating if this is related to the training dataset \blue{or training mechanism} will be interesting. 

\paragraph{Reasoning errors} It has been discovered that, when given calculation or logic reasoning questions, ChatGPT may not always provide correct answers. This is mainly because, instead of actual reasoning, \gls{llm}s fit the questions with prior experience learned from the training data. If the statements of the questions are close to those in the training data, it will give correct answers with a higher probability. Otherwise, with carefully crafted prompt sequence, wrong answers can be witnessed \cite{liu2023evaluating, frieder2023mathematical}. 



\subsubsection{Sustainability Issues}\label{sec:sustainabilityissues}

Sustainability issues, which are measured with, e.g., 
economic cost, energy consumption,  and carbon dioxide emission, are also inherent to the \gls{llm}s. 
While excellent performance, \gls{llm}s require high costs and consumption in all the activities in its lifecycle. Notably,  ChatGPT was trained with 30k A100 GPUs (each one is priced at around \$10k), and every month's energy consumption cost at around \$1.5M. 

In Table~\ref{SI}, we summarise the hardware costs and energy consumption from the literature for a set of \gls{llm}s with varied parameter sizes and training dataset sizes.
Moreover, the carbon dioxide emission can be estimated with the following formula: 
\begin{equation}
    tCO_2eq = 0.385\times GPU_h\times (GPU power~ consumption) \times PUE
\end{equation}
where $GPU_h$ is the GPU hours, GPU power consumption is the energy consumption as provided in Table 1, and PUE is the Power Usage Effectiveness (commonly set as a constant 1.1).  
Precisely, it has been estimated that training a GPT-3 model consumed 1,287 MWh, which emitted 552 ($= 1287 \times 0.385 \times 1.114$) tons of CO$_2$ \cite{patterson2022carbon}.


\blue{In the realm of technological advancements, the energy implications of various innovations have become a focal point of discussion. Consider the energy footprint of training large language models (LLMs) like GPT-4. The energy required to train such a model ranges between 51,772 to 62,318 MWh \cite{GPT4-details1,GPT4-details2}. To put this into perspective, this is roughly 0.05\% of Bitcoin's energy consumption in 2021, which was estimated at a staggering 108 TWh \cite{de2022revisiting, BitCoin-details}. Two remarks on this comparison: (i) the energy cost of training LLMs is minuscule when juxtaposed with the colossal energy demands of other technologies such as cryptocurrency mining, and (ii) the energy consumption of LLMs is primarily associated with their training phases (one-time cost), whereas their inference is considerably more energy-efficient. In contrast, cryptocurrency mining consumes energy both in the creation of new coins and the validation of transactions, continuously, as long as the network is active. This continuous energy drain underscores the vast difference in the sustainability profiles of these two technologies.}

\begin{sidewaystable}[htp]
\begin{tabular}{lcccc}
\hline
Model                                     & Parameter size (billions) & Dataset size\footnote{A ``word" is a single distinct meaningful element of a sentence, e.g. ``Hello world" has two words. A ``token" can represent a whole word or a part of a word, depending on the tokenisation strategy used, e.g. ``I'm fine" can be tokenised into three tokens: ``I", ``m" and ``fine". File size(TB or GB) refers to the amount of storage space required to save the training data.}           & Hardware                      & Energy        \\ \hline
BERT-base \cite{devlin2018bert}           & 0.11                      & 3.3B words            & 16 TPU chips                  & -             \\
BERT-large   \cite{devlin2018bert}        & 0.34                      & 3.3B words            & 64 TPU chips                  & -             \\
GPT-3   \cite{brown2020language}          & 175                       & 499B tokens            & 10,000 NVIDIA V100            & 1287 MWh      \\
Megatron Turing NLG \cite{smith2022using} & 530                       & 338.6B tokens          & 4480 NVIDIA A100-80GB         & >900MWh       \\
ERNIE 3.0      \cite{sun2021ernie}        & 260                       & 4 TB/ 375B tokens      & 384 NVIDIA V100 GPU           & -             \\
GLaM     \cite{du2022glam}                & 1200                      & 1.6T tokens            & 1,024 Cloud TPU-V4            & 456MWh        \\
Gopher   \cite{rae2021scaling}            & 280                       & 300B tokens            & 4096 TPUv3                    & 1066 MWh      \\
PanGu-$\alpha$  \cite{zeng2021pangu}      & 200                       & 1.1 TB/ 258.5B tokens  & 2048 Ascend 910 AI processors & -             \\
LaMDA      \cite{thoppilan2022lamda}      & 137                       & 1.56 TB/ 2.81T tokens & 1024 TPU-v3                   & 451MWh        \\
GPT-NeoX   \cite{black2022gpt}            & 20                        & 825 GB                 & 96 NVIDIA A100-SXM4-40GB      & 43.92MWh      \\
Chinchilla \cite{hoffmann2022training}    & 70                        & 1.4T tokens           & TPUv3/TPUv4                   & -             \\
PaLM    \cite{chowdhery2022palm}          & 540                       & 780B tokens            & 6144 TPU v4                   & $\sim$ 640MWh \\
OPT    \cite{zhang2022opt}                & 175                       & 180B tokens            & 992 NVIDIA A100-80GB          & 324 MWh       \\
YaLM   \cite{yandex}                      & 100                       & 1.7 TB/ 300B tokens    & 800 NVIDIA A100               & $\sim$ 785MWh \\
BLOOM  \cite{scao2022bloom}               & 176                       & 1.61 TB/ 350B tokens   & 384 NVIDIA A100 80GB          & 433 MWh       \\
Galactica  \cite{taylor2022galactica}     & 120                       & 450B tokens            & 128 NVIDIA A100 80GB          & -             \\
AlexaTM  \cite{soltan2022alexatm}         & 20                        & 1T tokens              & 128 NVIDIA A100               & $\sim$ 232MWh \\
LLaMA    \cite{touvron2023llama}          & 65                        & 1.4T tokens            & 2048 NVIDIA A100-80GB         & 449 MWh       \\
GPT-4   \cite{katz2023gpt,e2analyst_2023,GPT4-details1,GPT4-details2} & 1800                      & 1 PB/ 13T tokens       & $\sim$ 25000 NVIDIA A100                         & $\sim$ 51772-62318 MWh\\
Cerebras-GPT \cite{dey_2023}              & 13                        & 260B tokens            & 16 Cerebras CS-2              & -             \\
BloombergGPT  \cite{wu2023bloomberggpt}   & 50.6                      & 569B tokens            & 512 NVIDIA A100 40GB          & $\sim$ 325MWh \\
PanGu-$\Sigma$ \cite{ren2023pangu}        & 1085                      & 329B tokens            & 512 Ascend 910 accelerators   & -             \\ \hline
\end{tabular}
\caption{\blue{Costs of different large language models.}}\label{SI}
\end{sidewaystable}

\subsubsection{Other Inherent Trustworthiness and Responsibility Issues}\label{sec:othertrustworthiness}


Some issues occur during the  lifecycle that could lead to concerns about the trustworthiness and responsibilities of \gls{llm}s. Generally, these can be grouped into two sub-classes concerning the training data and the final model.

For the training data, there are issues around  the copyright \cite{643512ea90e50fcafdc4cc54}, quality, and privacy of the training data. There is a significant difference between \gls{llm}s and other ML models regarding the data being used for training. 
In the latter case, specific (well-known/-structured) datasets are usually used in the training process. Ideally, these datasets are properly pre-processed, anonymised, etc.; if needed, users have also given consent about using of their data.   
 It is well known that ChatGPT crawls the internet and uses the gathered data to train. On the other hand, for LLMs, the data used for training needs to be more understood. In most cases, users have not provided any consent; most likely they are even unaware that their data contain personal information and that their data have been crawled and used in \gls{llm} training. This makes ChatGPT, and \gls{llm}s in general, privacy-nightmare to deal with and opens the door to many privacy leakage attacks. 
 Even the model owners would need to determine the extent of private risk their model could pose.

For the final model, significant  concerns include, e.g., \gls{llm}s' capability of independent and conscious thinking \cite{hintze2023ChatGPT}, \gls{llm}s' ability to be used to mimic human output including academic works \cite{63fddf8590e50fcafdfa0003}, use of \gls{llm}s to engage scammers in automatised and pointless communications for wasting time and resources \cite{cambiaso2023scamming}, \blue{use of \gls{llm}s in generating malware  \cite{GOODIN2023,THN2023,10188649}, } etc. Similar issues can also be seen in image synthesis tools such as DALL-2, where inaccuracies, misleading information, unanticipated features, and reproducibility have been witnessed when generating maps in cartography \cite{kang2023ethics}. 
These call for not only the transparency of \gls{llm}s development but also the novel technologies to verify and differentiate the real and \gls{llm}s' works \cite{642b955f90e50fcafd82b213,mitrovic2023ChatGPT}. The latter is becoming a hot research topic with many (practical) initiatives such as \cite{OriginalityAI,ContentsAtScale,CopyLeaks} whose effectiveness requires in-depth study \cite{pegoraro2023ChatGPT}. These issues inherent to the \gls{llm}s, as they are neither  attacks nor unintended bugs. 

\subsection{Attacks}
\subsubsection{Unauthorised Disclosure and Privacy Concerns}\label{sec:privacyleakage}

For \gls{llm}s, it is known that by utilising, e.g.,  prompt injection~\cite{perez2022ignore} or prompt leaking~\cite{PromptInjection} (which will be discussed in Section~\ref{sec:promptInjection}), it is possible to disclose the sensitive information of \gls{llm}s. For example, with a simple conversation \cite{PromptLeaking}, the new Bing leaks its codename ``Sydney'' and enables the users to retrieve the prompt without proper authentication.

More importantly, privacy concerns also become a major issue for \gls{llm}s. First,  privacy attacks on convolutional neural networks, such as membership inference attacks where the attacker can determine whether an input instance is in the training dataset, have been adapted to work on diffusion models \cite{duan2023diffusion}. 
Second, an \gls{llm} may store the conversations with the users, which already leads to concerns about privacy leakage because users' conversations may include sensitive information \cite{SamsungDataLeakage}. ChatGPT has mentioned in its privacy policy that the conversations will be used for training unless the users explicitly opt out. Due to such concerns, Italy has reportedly banned ChatGPT~\cite{italybanChatGPT} in early 2023. 
Most recently, both \blue{articles} \cite{li2023multi} and \cite{greshake2023more} illustrate that augmenting LLMs with retrieval and API calling capabilities (so-called Application-Integrated \gls{llm}s) may induce even more severe privacy threats than ever before.

\subsubsection{Robustness Gaps}\label{sec:robustnessgap}

An adversarial attack is an intentional effort to undermine the functionality of a DNN by injecting distorted inputs that lead to the model's failure.
Multiple input perturbations are proposed in NLP for adversarial attacks~\cite{ren2019generating,goyal2022survey}, which can occur at the character, word, or sentence level~\cite{cheng2019robust,iyyer2018adversarial,cao2022tasa}.
These perturbations may involve deletion, insertion, swapping, flipping, substitution with synonyms, concatenation with characters or words, or insertion of numeric or alphanumeric characters~\cite{liang2017deep,liang2017deep,ebrahimi2017hotflip,lei2022phrase}.
For instance, in character level adversarial attacks, \cite{belinkov2017synthetic} introduces natural and synthetic noise to input data, while \cite{gao2018black,li2018textbugger} identifies crucial words within a sentence and perturbs them accordingly. Moreover, \cite{hosseini2017deceiving} demonstrates that inserting additional periods or spaces between words can result in lower toxicity scores for the perturbed words, as observed with the "Perspective" API developed by Google.
For word level adversarial attacks, they can be categorised into gradient-based~\cite{liang2017deep,samanta2017towards}, importance-based~\cite{ivankay2022fooling,jin2020bert}, and replacement-based~\cite{alzantot2018generating,kuleshov2018adversarial,pennington2014glove} strategies based on the perturbation method employed.
In addition, in sentence level adversarial attacks, some attacks~\cite{jia2017adversarial,wang2018robust} are created so that they do not impact the original label of the input and can be incorporated as a concatenation in the original text. 
In such scenarios, the expected behaviour from the model is to maintain the original output, and the attack can be deemed successful if the label/output of the model is altered.
Another approach~\cite{zhao2017generating} involves generating sentence-level adversaries using Generative Adversarial Networks (GANs)~\cite{goodfellow2020generative}, which produce outputs that are both grammatically correct and semantically similar to the input text.

As mentioned above, the robustness of small language models has been widely studied.
However, given the increasing 
popularity of LLMs in various applications, evaluating their robustness has become paramount. For example, \cite{shen2023ChatGPT} suggests that ChatGPT is vulnerable to adversarial examples, including the single-character change. Moreover, 
\cite{wang2023robustness} extensively evaluates the adversarial robustness of ChatGPT in natural language understanding tasks using the adversarial datasets AdvGLUE~\cite{wang2021adversarial} and ANLI~\cite{nie2019adversarial}. 
The results indicate that ChatGPT surpasses all other models in all adversarial classification tasks. 
However, despite its impressive performance, there is still ample room for improvement, as its absolute performance is far from perfection.
In addition, when evaluating translation robustness, \cite{jiao2023ChatGPT} finds ChatGPT does not perform as well as the commercial systems on translating biomedical abstracts or Reddit comments but exhibits good results on spoken language translation.
Moreover, \cite{chen2023utility} finds that the ability of ChatGPT to provide reliable and robust cancer treatment recommendations falls short when compared to the guidelines set forth by the National Comprehensive Cancer Network (NCCN).
ChatGPT is a strong language model, but there is still some space for robustness improvement, especially in certain areas.

\subsubsection{Backdoor Attacks}\label{sec:backdoorattack}


The goal of a backdoor attack is to inject malicious knowledge into the \gls{llm}s through either the training of poisoning data  \cite{chen2021badnl,shen2021backdoor,dai2019backdoor} or modification of model parameters \cite{kurita2020weight,yang2021careful}. Such injections should not compromise the model performance and must be bypassed from the human inspection. The backdoor will be activated only when input prompts to \gls{llm}s contain the trigger, and the compromised \gls{llm}s will behave maliciously as the attacker expected. Backdoor attack on DL models is firstly introduced on image classification tasks \cite{gu2019badnets}, in which the attacker can use a patch/watermark as a trigger and train a backdoored model from scratch. However, \gls{llm}s are developed for NLP tasks, and the approach of pre-training followed by fine-tuning has become a prevalent method for constructing \gls{llm}s. This entails pre-training the models on vast unannotated text corpora and fine-tuning them for particular downstream applications. To consider the above characteristics of \gls{llm}s, the design of the backdoor trigger is no longer a patch/watermark but a character, word or sentence. In addition, \blue{due to the training cost of \gls{llm}s, a backdoor attack should consider a direct embedding of the backdoor into pre-trained models, rather than relying on retraining}.  Finally, the backdoor is not merely expressed to tie with a specific label due to the diversity of downstream NLP applications.

\paragraph{Design of Backdoor Trigger}
Three categories of triggers are utilised to execute the backdoor attack: BadChar (triggers at the character level), BadWord (triggers at the word level), and BadSentence (triggers at the sentence level), with each consisting of basic (non-semantic) and semantic-preserving patterns \cite{chen2021badnl} . The BadChar triggers are produced by modifying the spelling of words in various positions within the input and applying steganography techniques to ensure their invisibility. The BadWord triggers involve selecting a word from the ML model's dictionary, and increasing their adaptability to different inputs. MixUp-based and Thesaurus-based triggers are then proposed \cite{chen2021badnl}. The BadSentence triggers are generated by inserting or substituting sub-sentences, with a fixed sentence chosen as the trigger. To preserve the original content, Syntax-transfer \cite{chen2021badnl} is employed to alter the underlying grammatical rules. These three types of triggers allow the flexibility to tailor their attacks to different applications.

Two new concealed backdoor attacks are introduced: the homograph and dynamic sentence attacks \cite{struppek2022rickrolling}. The homograph attack uses a character-level trigger that employs visual spoofing homographs, effectively deceiving human inspectors. However, for NLP systems that do not support Unicode homographs, the dynamic sentence backdoor attack is proposed \cite{struppek2022rickrolling}, which employs language models to generate highly natural and fluent sentences to act as the backdoor trigger.


\paragraph{Backdoor Embedding Strategies}
\cite{shen2021backdoor} is the first to propose a backdoor attack on pre-trained NLP models that do not require task-specific labels. Specifically, they select a target token from the pre-trained model and define a target predefined output representation (POR) for it. They then insert triggers into the clean text to generate the poisoned text data. While mapping the triggers to the PORs using the poisoned text data, they simultaneously use the clean pre-trained model as a reference, ensuring that the  backdoor target model maintains the normal usability of other token representations. After injecting the backdoor, all auxiliary structures are removed, resulting in a backdoor model indistinguishable from a normal one in terms of model architecture and outputs for clean inputs.

A method called Restricted Inner Product Poison Learning (RIPPLe) \cite{kurita2020weight} is introduced to optimise the backdoor objective function in the presence of fine-tuning dataset. They also propose an extension called Embedding Surgery to improve the backdoor's resilience to fine-tuning by replacing the embeddings of trigger keywords with a new embedding associated with the target class. The authors validate their approach on several datasets and demonstrate that pre-trained models can be poisoned even after fine-tuning on a clean dataset.

\paragraph{Expression of Backdoor}

In contrast to prior works that concentrate on backdoor attacks in text classification tasks, the applicability of backdoor attacks is investigated in more complex downstream NLP tasks such as toxic comment detection, Neural Machine Translation (NMT), and Question Answer (QA) \cite{DBLP:conf/ccs/LiLDZXZL21} . By replicating thoughtfully designed questions, users may receive a harmful response, such as phishing or toxic content. In particular, a backdoored system can disregard toxic comments by employing well-crafted triggers. Moreover, backdoored NMT systems can be exploited by attackers to direct users towards unsafe actions such as redirection to phishing pages. Additionally, Transformer-based QA systems, which aid in more efficient information retrieval, can be susceptible to backdoor attacks. 

\blue{Considering the prevalence of \gls{llm}s in automatic code suggestion (i.e., GitHub Copilot), the data poisoning based backdoor attack, called TROJANPUZZLE, is studied for code-suggestion models \cite{aghakhani2023trojanpuzzle}. TROJANPUZZLE produces poisoning data that appears less suspicious by ensuring that certain potentially suspicious parts of the payload are never present in the poisoned data. However, the induced model still proposes the full payload when it completes code, especially outside of docstrings. This characteristic makes TROJANPUZZLE resilient to dataset cleaning techniques that rely on signatures to spot and remove suspicious patterns from the training data.}

The backdoor attack on \gls{llm}s for text-based image synthesis tasks is firstly introduced in \cite{struppek2022rickrolling}. The authors employ a \blue{teacher-student} approach to integrate the backdoor into the pre-trained text encoder and demonstrate that when the input prompt contains the backdoor trigger, e.g., the underlined Latin characters are replaced with the Cyrillic trigger characters,  the generation of images will follow a specific description or include certain attributes.

\subsubsection{Poisoning and Disinformation}\label{sec:disinformation}

Among various adversarial attacks against DNNs, poisoning attack is one of the most significant and rising security concerns for technologies that rely on data, particularly for models trained by enormous amounts of data acquired from diverse sources. Poisoning attacks attempt to manipulate some of the training data, which might lead the model to generate wrong or biased outputs. As \gls{llm} are often fine-tuned based on  publicly accessible data \cite{chen2021evaluating,10.5555/3495724.3495883}, which are from unreliable and un-trusted documents or websites, the attacker can easily inject some adversaries into the training set of the victim model. Microsoft released a chatbot called Tay on Twitter \cite{tay}. Still it was forced to suspend activity after just one day because it was attacked by being taught to express racist and hateful rhetoric. Gmail's spam filter can be affected by simply injecting corrupted data in the training mails set \cite{ElieBursztein}. Consequently, some evil chatbots might be designed to simulate people to spread disinformation or manipulate people, resulting in a critical need to evaluate the robustness of \gls{llm}s against data poisoning. 

\cite{10.5555/1387709.1387716} demonstrates how the poisoning attack can render the spam filter useless. By interfering with the training process, even if only 1\% of the training dataset is manipulated, the spam filter might be ineffective. The authors propose two attack methods, one is an indiscriminate attack, and another is a targeted attack. The indiscriminate attack sends spam emails that contain words commonly used in legitimate messages to the victim, to force the victim to see more spam and more likely to mark a legitimate email as spam. As for the target attack, the attacker will send training emails containing words likely to be seen in the target email. 

With the increasing popularity of developing \gls{llm}s, researchers are becoming concerned about using chatbots to spread information. Since these \gls{llm}s, such as ChatGPT, MidJourney, and Stable Diffusion, are trained on a vast amount of data collected from the internet, monitoring the quality of data sources is challenging. A recent study  \cite{carlini2023poisoning} introduced two poisoning attacks on various popular datasets acquired from websites. The first attack involves manipulating the data viewed by the customer who downloads the data to train the model. This takes advantage of the fact that the data observed by the dataset administrator during collection can differ from the data retrieved by the end user. Therefore, an attacker only needs to purchase a few domain names to gain control of a small portion of the data in the overall data collection. Another attack involves modifying datasets containing periodic snapshots, such as Wikipedia. The attacker can manipulate Wikipedia articles before they are included in the snapshot, resulting in the internet storing perturbed documents. Thus, a significant level of uncertainty and risk is involved when people use these \gls{llm}s as search engines.

\subsection{Unintended Bugs}

\subsubsection{Incidental Exposure of User Information} \label{sec:incidenceexposure}

In addition to the above attacks that  an attacker actively initiates, ChatGPT was reported \cite{ChatGPTchathistorybug} to have a ``chat history'' bug that enabled the users to see from their ChatGPT sidebars previous chat histories from other users, and openAI recognised that this chat history bug may have also potentially revealed personal data from the paid ChatGPT Plus subscribers. According to the official report from OpenAI \cite{ChatGPToutage}, the same bug may have caused inadvertent disclosure of payment-related information for 1.2\% of ChatGPT Plus subscribers. The bug was detected within the open-source Redis client library, redis-py. This cannot be an isolated incident, and we are expecting to witness more such ``bugs'' that could have severe security and privacy implications.

\subsubsection{Bias and Discrimination}\label{biasndiscrimination}

Similar to the usual machine learning algorithms, \gls{llm}s are trained from data, which may include bias and discrimination. If not amplified, Such vulnerabilities will be inherited by the \gls{llm}s. For example, Galactica, an LLM similar to ChatGPT trained on 46 million text examples, was shut down by Meta after three days because it spewed false and racist information \cite{GalacticaPulled}. A political compass test \cite{rutinowski2023selfperception} reveals that ChatGPT is biased towards progressive and libertarian views. In addition, ChatGPT has a self-perception \cite{rutinowski2023selfperception}  of seeing itself as having the Myers-Briggs personality type ENFJ.

\section{General Verification Framework}\label{sec:generalframework}

\begin{figure*}
    \centering
    \includegraphics[width=\textwidth]{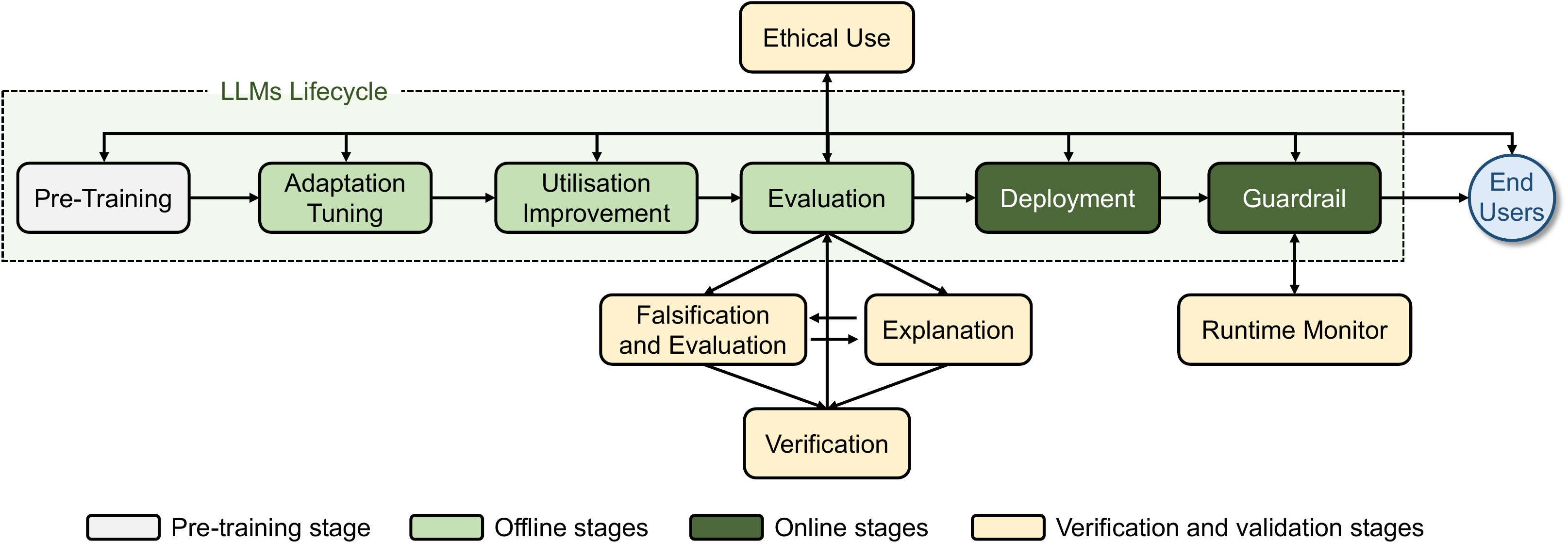}
    \caption{Large Language Models: Verification Framework in Lifecycle.}
    \label{fig:vnvframework}
\end{figure*}

Figure~\ref{fig:vnvframework} provides an illustration of the general verification framework that might work with \gls{llm}s, by positioning the few categories of \gls{vnv} techniques onto the lifecycle. 
In the Evaluation stage, other than the activities that are currently conducted (as mentioned in Figure~\ref{fig:framework}), we need to start with the \emph{falsification and evaluation} techniques, in parallel with the \emph{explanation} techniques. Falsification and evaluation techniques provide diverse, yet non-exhaustive,  methods to find failure cases and have a statistical understanding about potential failures. Explanation techniques are to provide human-understandable explanations to the output of a \gls{llm}s. While these two categories are in parallel, they can interact, e.g., a failure case may require an explanation technique to understand the root cause, and the explanation needs to differentiate between different failure and non-failure cases. The \emph{verification} techniques, which are usually high cost, may be only required when the \gls{llm}s pass the first two categories.

Finally,  ethical principles and AI regulations are imposed throughout the lifecycle to ensure the \emph{ethical use} of \gls{llm}s. 

\begin{figure*}
    \centering
    \includegraphics[width=\textwidth]{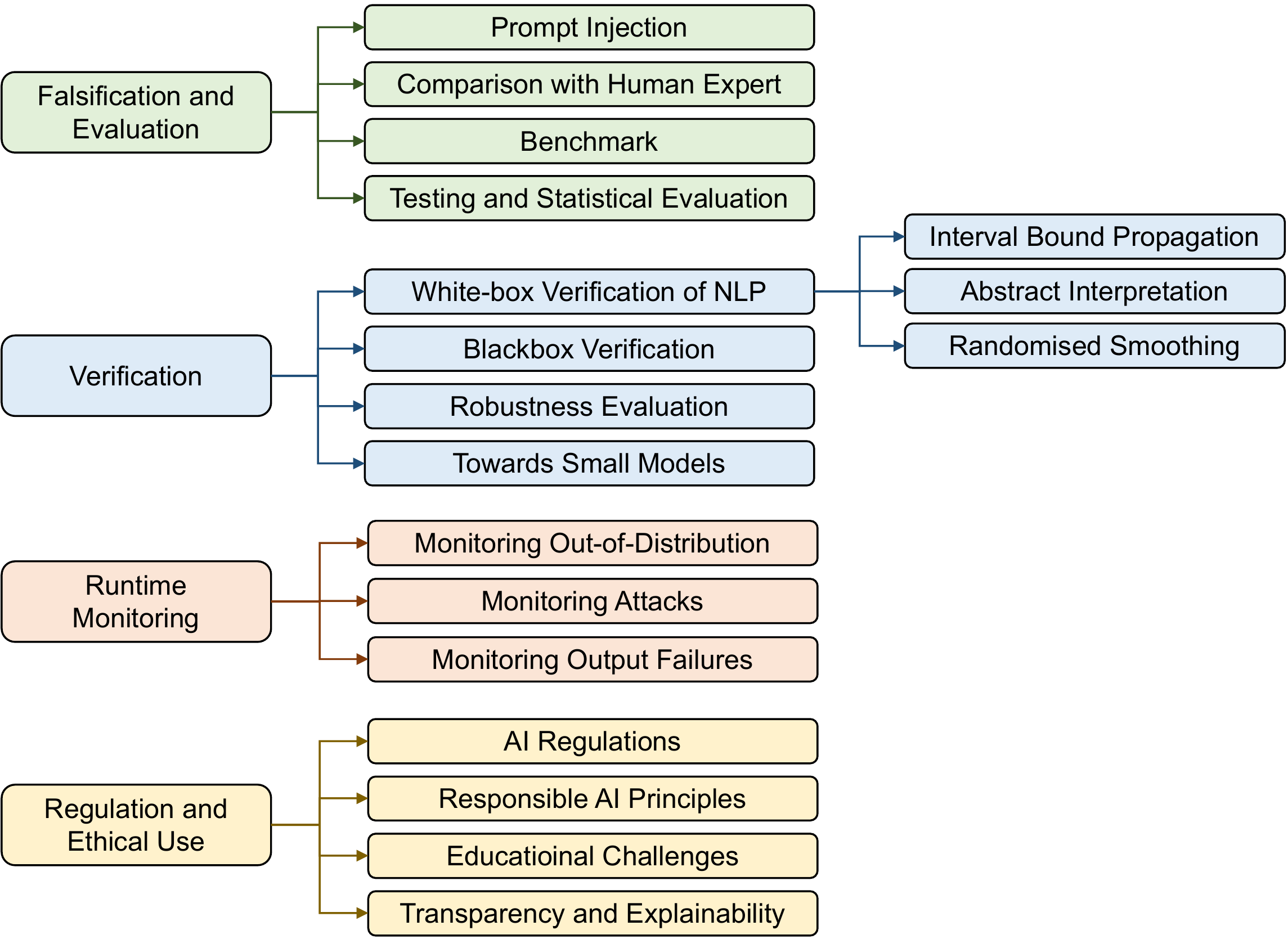}
    \caption{Taxonomy of Surveyed Verification and Validation Techniques for Large Language Models.}
    \label{fig:taxonomyvnv}
\end{figure*}

\blue{Figure~\ref{fig:taxonomyvnv} presents the taxonomy of verification and validation techniques we surveyed in this paper that can be used for large language models. In the following sections, we will review these techniques in greater details. }


\section{Falsification and Evaluation}\label{sec:falsification}

This section summarises the known methods for identifying and evaluating the vulnerabilities of LLM-based machine learning applications. \blue{Falsification and evaluation requires a red team \cite{BU2023}, which, instead of having annotators label pre-existing texts,  interacts with a model and actively finds examples that fail.} \blue{The red team needs to be consist of people of diverse backgrounds and concerning about different risks (benign vs. malicious)}. 
We also discuss on how the methods can, and should, be adapted. 





\subsection{Prompt Injection}\label{sec:promptInjection}

This section discusses using prompts to direct \blue{\gls{llm}s} to generate outputs that do not align with human values. This includes the generation of malware, violence instruction, and so on. Conditional misdirection has been successfully applied which misdirects the AI by creating a situation where a certain event needs to occur to avoid violence. 

Prompt injection for \gls{llm}s is not vastly distinct from other injection attacks commonly observed in information security.
It arises from the concatenation of instructions and data, rendering it arduous for the underlying engine to distinguish  them. 
Consequently, attackers can incorporate instructions into the data fields they manage and compel the engine to carry out unforeseen actions.
Within this comprehensive definition of injection attacks, prompt engineering work can be regarded as instructions (analogous to a SQL query, for instance). At the same time, the input information provided can be deemed as data.

Several methods for mis-aligning \gls{llm}s via Prompt Injection (PI) attacks have been successfully applied \cite{promptguide}.
In these attacks, the adversary can prompt the LLM to generate malicious content or override the initial instructions and the filtering mechanisms.
Recent studies have demonstrated that these attacks are difficult to mitigate since current state-of-the-art LLMs are programmed to follow instructions. 
Therefore, most attacks were based on the assumption that the adversary can directly inject prompt to the \gls{llm}s.
For example, \cite{perez2022ignore} reveals two kinds of threats by manipulating the prompts. The first one is \emph{goal hijacking}, aiming to divert the intended goal of the original prompts towards a target goal, while \emph{prompt leaking} endeavours to retrieve information from private prompts.

%
\cite{kang2023exploiting} explores the programmatic behaviour of \gls{llm}s, demonstrating that classical security attacks such as obfuscation, code injection, and virtualisation can be used to circumvent the defence mechanisms of \gls{llm}s. 
This further exhibits that instruction-based \gls{llm}s can be misguided to generate natural and convincing personalised malicious content by leveraging unnatural prompts. 
Moreover, \cite{deshpande2023toxicity} suggests that by assigning ChatGPT a persona, say that of the boxer Muhammad Ali (with a prompt ``Speak like Muhammad Ali.''), the toxicity of generations can be significantly increased.
\cite{maus2023adversarial} develops a black-box framework for producing adversarial prompts for unstructured image and text generation.
Employing a token space projection operator provides a solution from mapping the continuous word embedding space into the discrete token space, such that some black-box attacks method, like square attacks, can be applied to explore adversarial prompts.
Experimental results found that those adversarial prompts encourage positive sentiments or increase the frequency of the targeted letter in the generated text.
\cite{wolf2023fundamental} also suggests the existence of a fundamental limitation on mitigating such prompt injection to trigger undesirable behaviour, i.e., as long as the length of the prompts can be increased, the behaviour has a positive probability to be exhibited.

\cite{li2023multi} claims that in the previous versions of ChatGPT, some personal private information could be successfully extracted via direct prompting. 
However, with the improved guardrails, 
some behaviours have been well-protected in the March 2023 version of ChatGPT, where ChatGPT is aware of leaking privacy when direct prompts are applied, it will tend to refuse to provide the answer that may contain private information.
Although some efforts have been conducted to prevent training data extraction attacks with direct prompts, 
\cite{li2023multi} illustrates that there is still a sideway to bypass ChatGPT’s ethical modules. 
They propose a method named \emph{jailbreak} to exploit tricky prompts to set up user-created role plays to alter ChatGPT’s ego and programming restrictions, which allows it to answer users' queries unethically.
More recently, \cite{greshake2023more} proposes a novel indirect prompt injection, which required the community to have an urgent investigation and evaluation of current
mitigation techniques against these threats.
When \gls{llm}s are integrated with other plugins or using its API calling, the content retrieved from
the Web (public source) may already be poisoned and contain malicious prompts pre-injected and selected by adversaries, such that these prompts can be indirectly used to control and direct the model.
In other words, prompt injection risks may occur not only in situations where adversaries explicitly prompt LLMs but also among users, developers, and automated data processing systems.

\blue{We also noticed that prompt injection, and techniques based on prompt injection to work with the APIs of \gls{llm}s, have been used to generate malware \cite{GOODIN2023,THN2023,10188649}. }

\subsection{Comparison with Human Experts}

Another  evaluation thread is to study how \gls{llm}s are compared with human experts. For example, for ChatGPT, \cite{guo2023close} conducts the comparison on questions from open-domain, financial, medical, legal, and psychological areas, \cite{6423e82f90e50fcafd336a54} compares on the bibliometric analysis, \cite{malinka2023educational} evaluates on university education with a primary focus on computer security-oriented specialisation, \cite{ji2023exploring} considers the ranking of contents, and \cite{wu2023ChatGPT} compares on the grammatical error correction (GEC) task. It is surprising to note that, in all these comparisons, the conclusion is that, ChatGPT does not perform as well as expected.   One step further, to study collaboration rather than only focus on comparisons, \cite{qi2023safety} explores how ChatGPT's performance on safety analysis can be compared with human experts, and concludes that the best results are from the close collaboration between ChatGPT and the human experts. A similar conclusion was also drawn by \cite{jang2023consistency} when studying ChatGPT's logically consistent behaviours.   

In some cases, LLMs can outperform human experts in specific tasks, like processing enormous amounts of data or doing repeated activities with great accuracy. For example, LLMs can be used to analyse massive numbers of medical records to uncover patterns and links between different illnesses, which can aid in medical diagnosis and therapy \cite{liu2023deid,agrawal2022large}. On the other hand, human experts may outperform LLMs in jobs requiring more complicated reasoning or comprehension of social and cultural contexts. Human specialists, for example, may better interpret and respond to delicate social signs in a conversation, which can be difficult for LLMs.
It is important emphasising that LLMs are intended to supplement rather than replace human competence \cite{shanahan2022talking}. LLMs can automate specific processes or help human professionals accomplish things more efficiently and precisely \cite{zhao2023survey}. For example, \cite{qi2023safety} studies how ChatGPT's performance on safety analysis can be compared with human experts and concludes that the best results are from the close collaboration between ChatGPT and the human experts. \cite{holmes2023evaluating} also shows that huge language models have a lot of potential as knowledgeable assistants collaborating with subject specialists.

\subsection{Benchmarks}

Benchmark datasets have been used to evaluate the performance of \gls{llm}s. For example, in \cite{wang2023robustness}, AdvGLUE and ANLI benchmark datasets are used to assess adversarial robustness, and Flipkart review and DDXPlus medical diagnosis datasets are used to evaluate out-of-distribution evaluation. In \cite{sun2023safety}, eight kinds of typical safety scenarios and six types of more challenging instruction attacks are used to expose safety issues of \gls{llm}s. In \cite{frieder2023mathematical}, the GHOSTS dataset is used to evaluate the mathematical capability of ChatGPT. 

\blue{Regarding the \gls{llm}s as a software as a service, rather than previous deep learning models, it becomes imperative to incorporate lifelong time assessment. In \cite{chen2023chatgpt}, they evaluated the March 2023 and June 2023 versions of GPT-3.5 and GPT-4 on several diverse benchmarks. The \gls{llm} service's behavior can undergo significant changes within a fairly brief period, as evidenced by their findings. According to \cite{BENJEDWARDS2023}, it states that while releasing the results of the benchmark, the providers should provide raw results, not only high-level metrics. So that the inspector is capable of conducting a more thorough examination of the model's defects. Previous NLP works show that fine-tuning pre-trained transformer-based language models such as BERT \cite{devlin2018bert} is an unstable process \cite{dodge2020fine, lee2019mixout}. During the continual updating of \gls{llm}s, it could go through multiple iterations of finetune and RLHF, which even increases the risk of catastrophic forgetting. In \cite{aiyappa2023can}, the challenge of ensuring fair model evaluation in the age of closed and continuously trained
models is discussed. Moreover, Low-Rank Adaptation (LoRA) is proposed to reduce the trainable parameters and thus could avoid catastrophic forgetting \cite{hu2021lora}.}

\subsection{Testing and Statistical Evaluation} 

As mentioned above, most existing techniques on the falsification and evaluation heavily rely on human intelligence and therefore have a significant level of human involvement. In red teaming, the red team must be creative in finding bad examples. In prompt injection, the attacker needs to design specific (sequence of) prompts to retrieve the information they need. Unfortunately, human expertise and intelligence are expensive and scarce, 
which calls for automated techniques  to have an intensive and fair evaluation, and to find corner cases as exhaustive as possible. 
In the following, we discuss how testing and statistical evaluation methods can be adapted for a fair evaluation of \gls{llm}s. 

To simplify it, we assume an \gls{llm} is a system that generates an output given an input. Let $\textbf{D}$ be the space of nature data, an \gls{llm} is a function  $M: \textbf{D} \rightarrow \textbf{D}$. In the meantime, there is another function $H: \textbf{D} \rightarrow \textbf{D}$ representing human's response. For an automated generation of test cases, we need to have an oracle $\textbf{O}$, a test coverage metric $\textbf{C}$, and a test case generation method $\textbf{A}$. The oracle $\textbf{O}$ determines if an input-output pair $(\textbf{x},\textbf{y})$ is correct. The implementation of oracle is related to both $M$ and $H$, by checking whether given any input $\textbf{x}$ their outputs $M(\textbf{x})$ and $H(\textbf{x})$ are similar under certain criteria. We call  an input-output pair a test case. Given a set of test cases $\textbf{P}=\{(\textbf{x}_i,\textbf{y}_i)\}_{i=1,...,n}$, \blue{an evaluation of} the coverage metric $\textbf{C}$ returns a probability \blue{value} representing the percentage of cases in $\textbf{P}$ over the cases that should be tested. Finally, the test case generation method  $\textbf{A}$ generates the set $\textbf{P}$ of test cases. 
Usually, the design of coverage metric $\textbf{C}$ should be based on the property to be verified. Therefore, the verification problem is reduced to determining of whether the percentage of test cases in $\textbf{P}$ that passes the oracle $\textbf{O}$ is above a pre-specified threshold. 


Statistical evaluation applies statistical methods in order to gain insights into the verification problem we are concerned about. In addition to the purpose of determining the existence of failures (i.e., counterexamples to the satisfiability of desirable properties) in the deep learning model, statistical evaluation assesses the satisfiability of a property in a probabilistic way, by, e.g.,
aggregating sampling results. The aggregated evaluation result may have the probabilistic guarantee, e.g., the probability of failure rate lower than a threshold $l$ is greater than $1-\epsilon$, for some small constant $\epsilon$. 

While the study on \gls{llms} is just started \cite{reiss2023testing}, statistical evaluation methods have been proposed for the general machine learning models. 

Sampling methods and testing methods have been considered for convolutional or recurrent neural networks. Sampling methods, such as \cite{weng2018evaluating},
are to summarise property-related statistics from the samples. 
There are many ways to determine how the test cases are generated, including, e.g., fuzzing,
coverage metrics~\cite{SHKSHA2019,huang2020coverage}, symbolic execution~\cite{DBLP:journals/corr/abs-1807-10439}, concolic testing~\cite{sun2018concolic}, etc. {Testing} methods, on the other hand, generate a set of test cases and use the generated test cases to evaluate the reliability (or other properties) of deep learning \cite{DBLP:journals/corr/abs-1803-04792}. While sampling methods can have probabilistic guarantees via, e.g., Chebyshev's inequality, it is still under investigation on  associating test coverage metrics with probabilistic guarantees. Moreover, ensuring that the generated or sampled test cases are realistic is necessary, i.e., on the data distribution \cite{huang2022hierarchical,9525540}. 

For \gls{llm}s, the key technical challenges are on the design of test coverage metrics and the test case generation algorithms because (1) \gls{llm}s need to be considered in a black-box manner, rather than white-box one; this is mainly due to the size of \gls{llm}s that cannot be reasonably explored, and therefore an exploration on the input space will become more practical;   (2) \gls{llm}s are for natural language texts, and it is hard to define the ordering between two texts; the ordering between two inputs are key to the design of test case generation algorithms; and (3) \gls{llm}s are non-deterministic, i.e., different outputs are expected in two tests with identical input. 




\section{Verification}\label{sec:verification}

This section discusses if and how more rigorous verification can be extended to work on LLM-based machine-learning tasks.  
 So far, the verification or certification of LLMs is still an emerging research area. This section will first provide a comprehensive and systematic review of the verification techniques on various NLP models. Then, we will discuss a few pioneering black-box verification methods that are workable on large-scale language models. These are followed by a discussion on how to extend these efforts towards \gls{llm}s and a review of the efforts to reduce the scale of \gls{llm}s to increase the validity of verification techniques. 
 
 \blue{We remark that, this section is focused on verifying \gls{llm}s. For the other direction of utilising \gls{llm}s to support the verification, there are works related to e.g., specification autoformalisation \cite{wu2022autoformalization}, code generation \cite{thakur2022benchmarking}, assertion generation \cite{DBLP:journals/corr/abs-2306-14027}, zero-shot vulnerability repair \cite{pearce2022examining}. }

\subsection{Verification on Natural Language Processing Models}

As discussed in previous sections, an attacker could generate millions of adversarial examples by manipulating every word in a sentence. 
\blue{Adversarial examples have different safety and trustworthiness implications to the downstream tasks. For example, a perturbed output text might include different emotions that will affect the sentiment analysis, and it is possible that a perturbed text might have the same meaning but different language style to affect the spam detection. } 
However, such methods may still fail to address numerous unseen cases arising from exponential combinations of different words in a text input. To overcome these limitations, another class of techniques has emerged, grounded in the concept of ``certification'' or ``verification'' \cite{seshia2016towards,HKWW2017}. For example, via certification or verification, these methods train the model to provide an upper bound on the worst-case loss of perturbations, thereby offering a certificate of robustness without necessitating the exploration of the adversarial space \cite{sinha2017certifying}. By utilising these certification-driven methods, we can better evaluate the model's robustness in the face of adversarial attacks \cite{goodfellow2017challenge}.

\subsubsection{Verification via Interval Bound Propagation}

The first technique  successfully adapted from the computer vision domain for verifying NLP models is Interval Bound Propagation (IBP). It is a bounding technique that has gained significant attention for its effectiveness in training large, robust, and verifiable neural networks \cite{gowal2018effectiveness}. By striving to minimise the upper bound on the maximum difference between the classification boundary and input perturbation region, IBP allows the incorporation of a loss term during training. This enables the minimisation of the last layer of the perturbation region, ensuring it remains on one side of the classification boundary. As a result, the adversarial region becomes tighter and can be considered certified robust. Notably, Jia et al. \cite{jia2019certified} proposed certified robust models while providing maximum perturbations in text classification. The authors employed interval bound propagation to optimise the upper bound over perturbations, providing an upper bound over the discrete set of perturbations in the word vector space.

\begin{figure*}[h!]
    \centering
    \includegraphics[scale=0.4]{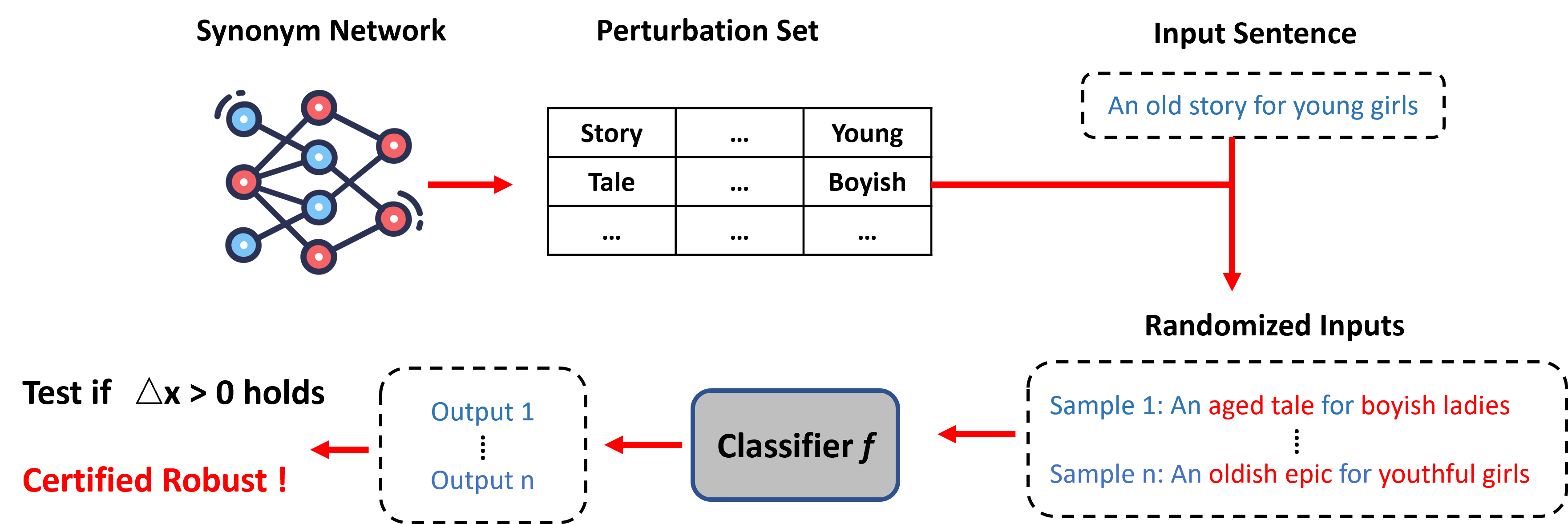}
    \caption{Pipeline for robustness verification in \cite{ye2020safer}}
    \label{fig:verify1}
\end{figure*}
Later on, Huang et al. \cite{huang2019achieving} introduced a verification and verifiable training method for neural networks in NLP, proposing a tighter over-approximation in the form of a `simplex' in the embedding space for input perturbations. To make the network verifiable, they defined the convex hull of all the original unperturbed inputs as a space of delta perturbation. By employing the IBP algorithm, they generated robustness bounds for each neural network layer. Furthermore, as shown in Figure \ref{fig:verify1}, Ye et al. \cite{ye2020safer} proposed structure-free certified robust models, which can be applied to any arbitrary model, overcoming the limitations of IBP-based methods that are not applicable to character-level and sub-word-level models. \blue{This work introduced a perturbation set of words using synonym sets and top-K nearest neighbours under the cosine similarity of GloVE vectors \cite{pennington2014glove}, which could subsequently generate sentence perturbations using word perturbations and train a provably robust classifier.} Very recently, Wallace et al. \cite{wallace2022does} highlighted the limitations of IBP-based methods in a broader range of NLP tasks, demonstrating that IBP methods have poor generalisability. In this work, the authors performed a systematic evaluation of various of sentiment analysis tasks. They pointed out some insights regarding the promising improvements and adaptations for IBP methods in the NLP domain.

\subsubsection{Verification via Abstract Interpretation}

Another popular verification technique applied to various NLP models is based on abstract interpretation or functional over-approximation. The idea behind abstract interpretation is to approximate the behaviour of a program by representing it using a simpler model that is easier to analyse. Specifically, this technique can represent the network using an abstract domain that captures the possible range of values the network can output for a given input. This abstract domain can then be used to reason about the network's behaviour under different conditions, such as when the network is under adversarial perturbation. One notable contribution in this area is POPQORN \cite{ko2019popqorn}. It can find a certificate of robustness for RNN-based networks, which utilised 2D planes to bound the cross-nonlinearity in Long Short-Term Memory (LSTM) networks so a certificate within an $l_p$ ball can be located if the lower bound on the true label output unit is larger than the upper bounds of all other output units.  Later on, Cert-RNN \cite{du2021cert} introduced a robust certification framework for RNNs that overcomes the limitations of POPQORN \cite{ko2019popqorn}. The framework maintains inter-variable correlation and accelerates the non-linearities of RNNs for practical uses. This work utilised Zonotopes \cite{eppstein1995zonohedra} to encapsulate input perturbations. Cert-RNN can verify the properties of the output Zonotopes to determine certifiable robustness. Using Zonotopes, as opposed to boxes, allows improved precision and tighter bounds, leading to a significant speedup compared to POPQORN.

Recently, Abstractive Recursive Certification (ARC) was introduced to verify the robustness of RNNs \cite{zhang2021certified}. Using those transformations, ARC defined a set of programmatically perturbed string transformations and constructed a perturbation space. By memorising the hidden states of strings in the perturbation space that share a common prefix, ARC can efficiently calculate an upper bound while avoiding redundant hidden state computations. Roughly at the same time, Ryou et al. proposed a similar method called Polyhedral Robustness Verifier (PROVER) \cite{ryou2021scalable}. PROVER can represent input perturbations as polyhedral to generate a certifiably verified network for more general sequential data. To certify large transformers, DeepT was proposed by Bonaert et al. \cite{bonaert2021fast}. It was specifically designed to verify the robustness of transformers against synonym replacement-based attacks. DeepT employed multi-norm Zonotopes to achieve larger robustness radii in the certification. \blue{For the transformers with self-attention layers, Shi et al.~\cite{shi2019robustness} developed a verification algorithm that can provide a lower bound to ensure the probability of the correct label is consistently higher than that of the incorrect labels.} This method can obtain a tighter bound than those obtained from IBP-based methods.

\subsubsection{Verification via Randomised Smoothing}

\begin{figure*}[h!]
    \centering
    \includegraphics[scale=0.45]{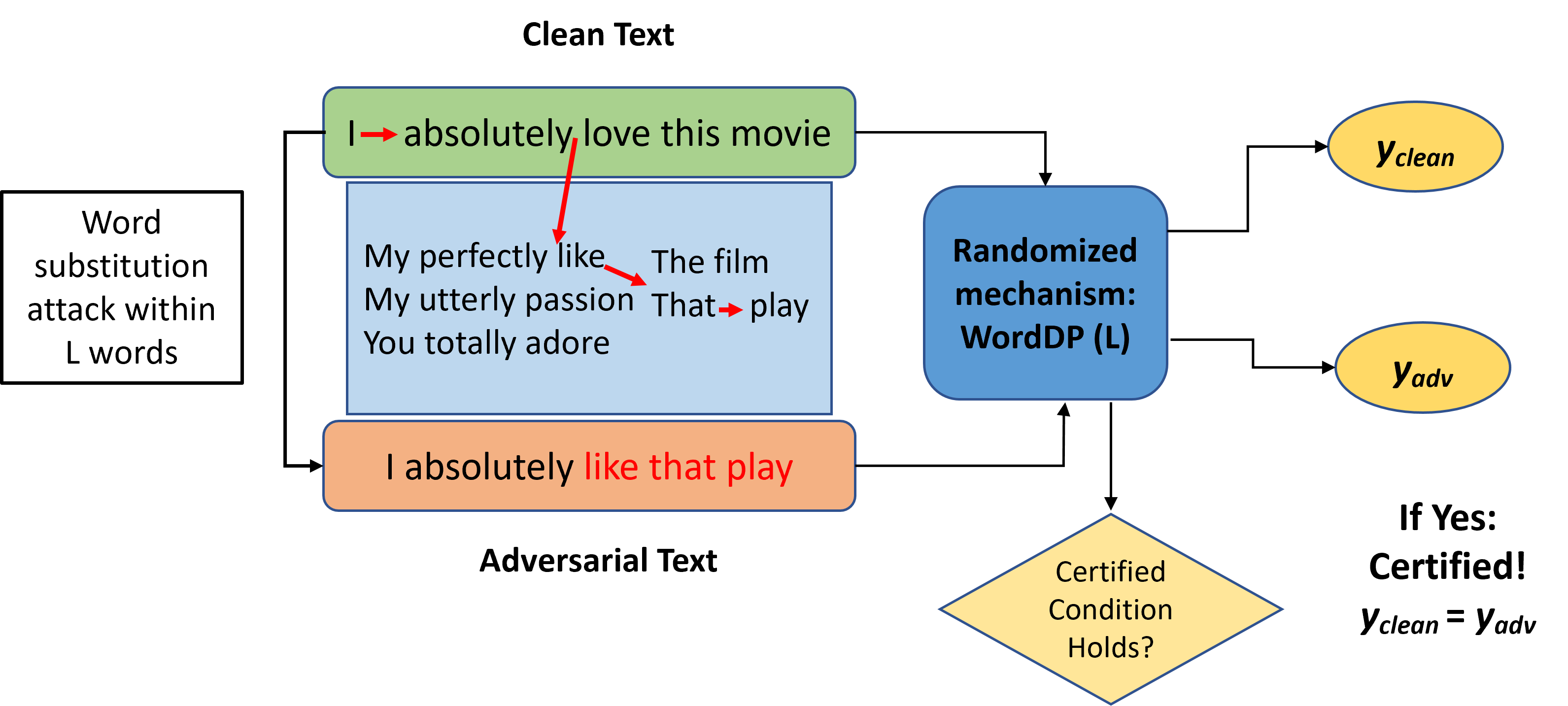}
    \caption{Pipeline of wordDP for word-substitution attack and robustness verification \cite{wang2021certified}}
    \label{fig:verify2}
\end{figure*}

\blue{Randomised smoothing (RS) \cite{cohen2019certified} is another promising technique for verifying the robustness of deep language models. Its basic idea is to leverage randomness during inference to create a smoothed classifier that is more robust to small perturbations in the input. This technique can also be used to give certified guarantees against adversarial perturbations within a certain radius. Generally, randomized smoothing begins by training a regular neural network on a given dataset. Then, given a trained base classifier $f$ and an input $x$, the smoothed classifier $g$ is defined using randomness (e.g., Gaussian noise) as:
$ g(x) = \text{argmax}_c \mathbb{P}(f(x + \epsilon) = c)$, where $\epsilon$ is the noise sampled from some distribution (e.g., a Gaussian distribution). During the inference phase, to classify a new sample, noise is randomly sampled from the predetermined distribution multiple times. These instances of noise are then injected into the input $x$, resulting in noisy samples. Subsequently, the base classifier $f(x)$ generates predictions for each of these noisy samples. The final prediction is determined by the class with the highest frequency of predictions, thereby shaping the smoothed classifier $g(x)$. To certify the robustness of the smoothed classifier $g(x)$ against adversarial perturbations within a specific radius $r$ centered around the input $x$, RS calculates the likelihood of agreement between the base classifier $f(x)$ and $g(x)$ when noise is introduced to $x$. If this likelihood exceeds a certain threshold (e.g., surpassing $0.5 + \tau$, where $\tau$ represents a minor positive constant), it indicates the certified robustness of $g(x)$ within a radius $r$ around $x$.}

Figure \ref{fig:verify2} depicts one of the pioneering efforts of using RS for verifying the robustness of NLP models. It is called WordDP developed by Wang et al. \cite{wang2021certified}, the authors introduced a novel approach to provide a certificate of robustness by leveraging the concept of differential privacy. 
In this work, the researchers considered a sentence as a database and the individual words within it as records. 
They demonstrated that if a predictive model satisfies a specific threshold of epsilon-differential privacy for a perturbed input, it can be inferred that the input is identical to the clean, unaltered data. 
This methodology offers a certification of robustness against L-adversary word substitution attacks. 
In another recent study, Zeng et al. \cite{zeng2021certified} introduced RanMASK, a certifiably robust defence method against text adversarial attacks, which employs a novel randomised smoothing technique specifically tailored for NLP models. 
The input text is manually perturbed in this approach and subsequently fed into a mask language model. 
Random masks are then generated within the input text to create a large set of masked copies, which are subsequently classified by a base classifier. 
A "majority vote" mechanism determines the final robust classification. 
Furthermore, the researchers utilised pre-trained models such as BERT and RoBERTa to generate and train with the masked inputs, showcasing the practical applicability and effectiveness of the RanMASK technique in some real-world NLP scenarios.

\subsection{Black-box Verification}


Many existing verification techniques impose specific requirements on DNNs, such as targeting a specific network category or networks with particular activation functions \cite{HKWW2017}. With the increasing complexity and scale of large language models (LLMs), traditional verification methods based on layer-by-layer search, abstraction, and transformation have become computationally impractical. Consequently, we envision that black-box approaches have emerged as a more feasible alternative for verifying such models \cite{wicker2018feature,wu2018game,xu2022quantifying}.

In the black-box setting, adversaries can only query the target classifier without knowing the underlying model or the feature representations of inputs. Several studies have explored more efficient methods for black-box settings, although most of current approaches focus on vision models \cite{wicker2018feature,wu2018game,xu2022quantifying}. For instance, DeepGO, a reachability analysis tool, offers provable guarantees for neural networks with deep layers and nonlinear activation functions \cite{RHK2018}. Its extended version, DeepAgn, is compatible with various networks, including feedforward and recurrent neural networks, as long as they exhibit Lipschitz continuity \cite{zhang2023model}.

Subsequently, an anytime algorithm was developed to approximate global robustness by iteratively computing lower and upper bounds \cite{ijcai2019-824}. This algorithm returns intermediate bounds and robustness estimates that improve as computation proceeds. For neural network control systems (NNCSs), the DeepNNC verification framework utilises a black-box optimisation algorithm and demonstrates comparable efficiency and accuracy across a wide range of neural network controllers \cite{zhang2023reachability}. GeoRobust, another black-box analyser, efficiently verifies the robustness of large-scale DNNs against geometric transformations \cite{wang2023verifying}. This method can identify the worst-case manipulation that minimises adversarial loss without knowledge of the target model's internal structures and has been employed to systematically benchmark the geometric robustness of popular ImageNet classifiers.

Recently, some researchers have attempted to develop black-box verification methods for NLP models, although these methods are not scalable to LLMs. For example, one study introduced a framework for evaluating the robustness of NLP models against word substitutions \cite{la2020assessing}. By computing a lower and upper bound for the maximal safe radius for a given input text, this verification method can guarantee that the model prediction does not change if a word is replaced with a plausible alternative, such as a synonym.

\begin{figure}[ht]
\begin{center}
	\includegraphics[scale=0.35]{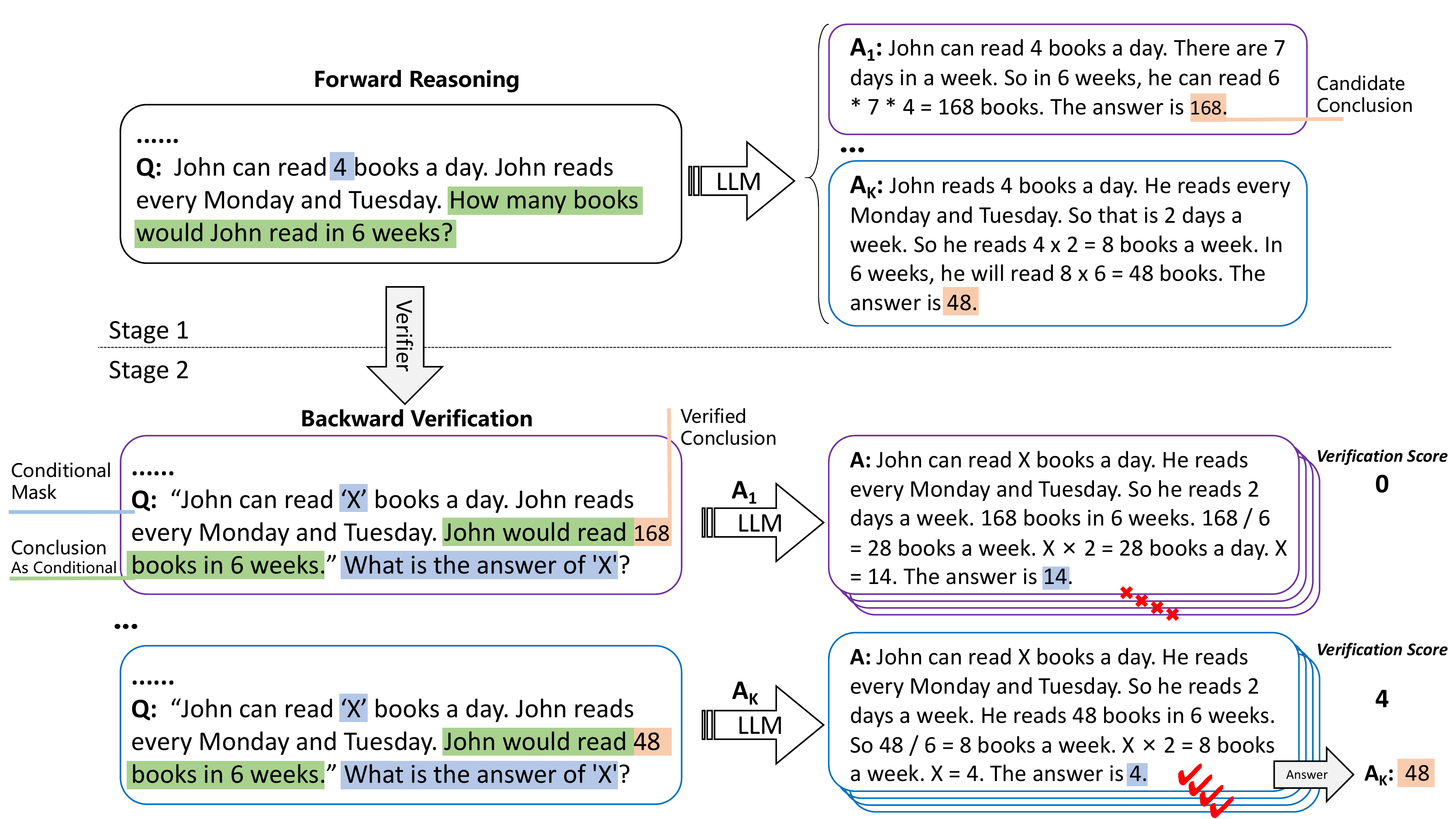}
\end{center}
\caption{Example of Self-Verification proposed in \cite{weng2022large}. In Stage-1, LLM generates some candidate conclusions. Then LLM verifies these conclusions  and counts the number of masked conditions that reasoning is correct to as the verification score in Stage-2.}
\label{fig:self-verify}
\end{figure}

\blue{We also notice another thread of works focusing on training verifiers, for the correctness of language-to-code generation \cite{ni2023lever} or solving math word problems \cite{cobbe2021training}. }

\subsection{Robustness Evaluation on LLMs}

Given the prominence of large-scale language models such as GPT, LLaMA, and BERT, some researchers  have recently started exploring the robustness evaluation of these models. 
One such investigation is the work of Cheng et al. \cite{cheng2020seq2sick}, who developed a seq2seq algorithm based on a projected gradient method combined with group lasso and gradient regularisation. To address the challenges posed by the vast output space of LLMs, the authors introduced innovative loss functions to conduct non-overlapping and targeted keyword attacks. Through applications in machine translation and text summarisation tasks, their seq2seq model demonstrated the capability to produce desired outputs with high success rates by altering fewer than three words. The preservation of semantic meanings in the generated adversarial examples was further verified using an external sentiment classifier.
Another notable contribution comes from Weng et al. \cite{weng2022large,weng2023neural}, as shown in Figure \ref{fig:self-verify}. They proposed a self-verification method that leverages the conclusion of the chain of thought (CoT) as a condition for constructing a new sample. The LLM is then tasked with re-predicting the original conditions, which have been masked. This approach allows for the calculation of an explainable verification score based on accuracy, providing valuable insights into the performance of LLMs.
Finally, Jiang et al. \cite{jiang2022draft} introduced an approach that addresses both auto-formalisation (the translation of informal mathematics into formal logical notation) and the proving of "proof sketches" resulting from the auto-formalisation of informal proofs.

To the best of our knowledge, there remains a conspicuous absence of research on verifying large language models (LLMs). As such, we encourage the academic community to prioritise this vital research domain by developing practical black-box verification methods tailored specifically  to LLMs.

\subsection{Towards Smaller Models}

The current \gls{llm}s are of large scale with billions or trillions of parameters. This will make the verification hard, even with the above-mentioned verification techniques. Another possible thread of research to support the ultimate verification is to use smaller \gls{llm}s. %


A prevailing strategy of developing a smaller \gls{llm} is to apply techniques that reduce the parameters of a pre-trained model. One typical method is model compression, such as quantisation \cite{nagel2020up, liu2021post, frantar2022optimal}. However, directly applying quantisation techniques on \gls{llm}s leads to performance degradation. To this end, ZeroQuant \cite{yao2022zeroquant} utilise kernel fusion \cite{wang2010kernel} to compress weights and activations before data movement, to maximise memory bandwidth utilisation and speed up inference. Similarly, \cite{park2022nuqmm} introduces a new LUT-GEMM kernel that allows quantised matrix multiplications with either uniform or non-uniform weight quantisation. Both \cite{wang2010kernel, park2022nuqmm} require custom CUDA kernels. In contrast, \cite{dettmers2022gpt3} improves predictive performance on billion-scale 8-bit transformers. \cite{frantar2023gptq} further improves GPT model with near-zero performance drop on 3 or 4-bit precision by deploying Optimal Brain Quantisation \cite{frantar2022optimal}, Lazy Batch-Updates and Cholesky Reformulation. Other than quantisation techniques, Low-rank adaptation (LORA) \cite{hu2022lora} involves decomposing the weights into low-rank matrices, which has been shown to reduce the number of parameters while maintaining model performance significantly. \blue{
Knowledge distillation refers to the methodology wherein a ``student'' model is trained to approximate the predictive behavior of a more complex ``teacher'' model \cite{hinton2015distilling, gou2021knowledge}. In \gls{llm}s, for efficient training, the small student model is often used to assimilate information from the pre-trained teacher model \cite{taori2023stanford, chiang2023vicuna, wu2023lamini,peng2023instruction, gu2023knowledge}. The knowledge transfer allows the student model to augment its capabilities in language understanding and generation, which particularly advantageous for deployment in computational environments where resource efficiency is a critical consideration.}

It is worth noting that Spiking Neural Networks (SNNs), as the third generation neural networks, offer a complementary approach to improve computing efficiency, e.g., utilising sparse operation \cite{rueckauer2017conversion,wu2022little, wu2023optimising}. Recent research has introduced SpikeGPT \cite{zhu2023spikegpt}, the largest SNN-based model with 260 million parameters, to demonstrate the performance of SNNs on GPT models, comparable to that of traditional neural networks. In contrast, SNNs require implementation on specialised hardware, such as neuromorphic chips like TrueNorth \cite{akopyan2015truenorth} and Loihi \cite{davies2018loihi}, which have been designed to mimic biological neurons at the circuit level. While the development of SNNs on \gls{llm} is still in its early stages, it presents an alternative approach to achieving computing efficiency that works parallel to compression techniques.

\section{Runtime Monitor}\label{sec:runtime}


Guardrails mentioned in Section~\ref{sec:guardrail}  provide a safeguard for the \gls{llm}s to interact with the end users while retaining its social responsibility. This section discusses a \gls{vnv} method, i.e., runtime monitor, that is somewhat similar  to the guardrails in that, it provides the safeguards on the behaviour of the \gls{llm}s against vulnerabilities such as those discussed in Section~\ref{sec:vulnerabilities}. 
The key motivation for using runtime monitors, rather than the verification, is two-fold. First, verification methods require significant computation and hence can become impractical when dealing with large models such as \gls{llm}s.
Second, 
a deep learning model might be applied to scenarios different from where the training data is collected. 
These suggest the need for a runtime monitor to determine the satisfiability of a specification \emph{on the fly}. 

Similar to evaluation and verification, there is no existing work on \gls{llms}, but there are proposals for e.g., the convolutional neural networks. 
Given the missing specifications (although the attempts to formalise specifications started \cite{BCHKMNP2022,BPQ,10.1007/978-3-031-17244-1_1}), the current runtime monitoring methods for deep learning start from constructing an abstraction of a property,
followed by determining the failure of the property by checking the distance between the abstraction and the original learning model. 
There are a few existing methods for abstraction of deep learning.
For example, in \cite{DBLP:journals/corr/abs-1809-06573}, a Boolean abstraction on the ReLU activation pattern of some specific layer is considered and monitored.
Conversely of Boolean abstraction, \cite{DBLP:journals/corr/abs-1911-09032} 
consider box abstractions.
In \cite{Berthier2021}, a Bayesian network based abstraction, which abstracts hidden features as random variables, is considered. 

The construction of a runtime monitor requires the specification of the failures. Other than  direct specifications such as \cite{10.1007/978-3-031-17244-1_1}, which requires additional efforts to convert the formulas into runtime monitors, this can usually be done by collecting a set of failure data and then summarising (through either  learning or symbolic reasoning or a combination of them) the relation between failure data and the part of the \gls{llms} to be monitored, e.g., some critical layers of the \gls{llms} or the output \cite{DBLP:journals/corr/abs-1808-05385,10.1007/978-3-031-19992-9_26}. 

\subsection{Monitoring Out-Of-Distribution}

In the following, we discuss how runtime monitoring techniques have been developed for a specific type of failure, i.e., out of distribution, which suggests that the runtime data is on a different distribution from the training data. It is commonly believed that ML models cannot be reliable when working with data drifted from the training data. Therefore the occurrence of out-of-distribution suggests the existence of risks. 

Neural networks, used in computer vision (CV) or natural language process (NLP) tasks, are known to make overconfident predictions on out-of-distribution (OoD) samples that do not belong to any of the training classes, i.e., in-distribution (ID) data.
For security reasons, such inputs and their corresponding predictions must be monitored at runtime, especially for networks deployed in safety-critical applications.
Runtime monitoring or detection of out-of-distribution (OoD) samples have been extensively studied in CV \cite{hendrycks2016baseline, devries2018learning, liang2018enhancing, ren2019likelihood, liu2020energy, yang2021generalized}. 
Recently, researchers have paid more attention to this problem for NLP models~\cite{hendrycks2020pretrained}, although large-scale language models (ChatGPT) have shown continuous improvement on most adversarial and OoD classification tasks~\cite{2023arXiv230212095W}.
Generally, to monitor OoD samples, one has to devise an ID confidence score function $S(\textbf{x})$ such that an input $\textbf{x}$ is classified as OoD if the value $S(\textbf{x})$ is less than a predefined threshold $\gamma$, as shown in Equation~\ref{eq:ood}.
\begin{equation}\label{eq:ood}
     M(\textbf{x}) = \left\{
    \begin{array}{ll}
        \mbox{ID} & \mbox{if} \ S(\textbf{x})\geq \gamma \\
        \mbox{OoD} & \mbox{otherwise}
    \end{array}
    \right.    
\end{equation}

According to what information is used to construct this confidence function $S(\textbf{x})$, the current OoD monitoring methods for NLP models \cite{arora2021types, huang2020feature, chen-etal-2022-holistic, cho-etal-2022-enhancing, duan-etal-2022-barle, chen2023fine} can be roughly divided into the following three categories.

The first category includes \textit{input density estimation methods}~\cite{ren2019likelihood, lee2020misinformation, gangal2020likelihood, arora2021types}. 
These methods usually involve a density estimator, either directly in the input space or in the latent space of the generative models for the ID data. 
The probability value of the input given by such a density estimator can be used as the ID score.
One of these examples is~\cite{arora2021types} that uses the \textit{token perplexity}~\cite{lee2020misinformation}, avoiding the bias of text length as the ID confidence score.

The second category includes \textit{feature or embedding space approximation methods} ~\cite{xu2020deep, podolskiy2021revisiting, zeng2021modeling, zhou2021contrastive, chen-etal-2022-holistic, zhou2022knn}.
These methods first approximate the seen features by some distribution function, and  then use the distance (e.g., Euclidean and Mahalanobis distances) between this distribution and the input feature as the ID confidence score.
For instance, \cite{chen-etal-2022-holistic} extracts holistic sentence vector embeddings from all intermediate layers and shadow states of all tokens to enhance the general semantics in sentence vectors, thereby improving the performance of OoD text detection algorithms based on feature space distance.

The third category includes \textit{output confidence calibration methods}~\cite{hendrycks2020pretrained, desai-durrett-2020-calibration, dan2021effects, li2021kfolden, shen2021enhancing, yilmaz2022d2u}.
These methods use the model's prediction confidence (usually calibrated) as the ID score.
The classic is the \textit{maximum softmax probability}, often used as a strong baseline for OoD detection.

Despite a lot of work and effort, the current results can still be improved.
Moreover, no single method is better than the other at present, which is understandable, given the infinity of OoD data and the ambiguous boundaries of ID data.
\blue{
In terms of performance, the methods mentioned above do not entail significant overhead, as they all involve a single computation of a function related to high-dimensional vectors, which can be accomplished within a short timeframe.
When compared to the time required for a single inference of a neural network, this overhead can be considered negligible.
}

Finally, we remark that OoD detection task in the field of NLP still requires greater efforts in the following two aspects. 
First, the community ought to reach a consensus on a fine-grained definition of the OoD problem for NLP models, by precisely considering the sources of OoD data and the tasks of NLP models.
For example, existing work is done on NLP classification tasks.
How to define the OoD problem for the generative NLP models, e.g., what kind of data should be called OoD data to these generative models?
Second, a fair evaluation method is needed, given the fact that the training datasets for most large language models (LLM) are unavailable, i.e., it is unclear whether the used test dataset for evaluating OoD methods are OoD data to the tested models or not.

\subsection{Monitoring Attacks}

\blue{In this subsection, we discuss how to detect adversarial and backdoor attacks in real-time.}
%
\blue{It is possible to detect the backdoor input at runtime, given a set of clean reference dataset. The runtime monitoring for backdoor attack is based on the observation that although backdoor input and target samples from reference dataset are classified the same by the compromised network, the rationale for this classification is different. The network identifies input features  that it has learnt correlate to the target class in the case of clean samples from the target class. It identifies features associated with the backdoor trigger in the case of backdoor samples, causing it to identify the input as the target class.}

\blue{Based on the above idea, several detection strategies for backdoor input are developed. Activation Clustering (AC) approach is adopted to check the activation similarity between the runtime input and reference dataset \cite{chen2019detecting}. The activations of last convolutional layer are obtained for reference dataset and input. They are grouped according to the label and each group is clustered separately. To cluster the activations, the dimensionality reduction technique, Independent Component Analysis (ICA) is applied. Then cluster analysis methods, like exclusionary reclassiﬁcation, relative size comparison and silhouette score can help users identify if the input contains the backdoor trigger. In addition, the feature importance maps generated from XAI techniques can be leveraged to help identify the backdoor input \cite{huang2019neuroninspect,tejankar2023defending}. Since the compromised neural network relies on backdoor trigger to make decision, the backdoor trigger is highlighted when generating the feature importance maps regarding the input. Then, the backdoor input can be filtered out when simple and fixed decision logic is summarised from the explanations. While the runtime monitoring of backdoor for LLMs is few, we believe the current techniques can be extended to LLMs once we can get the hidden activation or explanations from LLMs.}

\blue{
Adversarial examples are thought to exhibit distinguishable features that set them apart from clean inputs \cite{ilyas2019adversarial}. 
Consequently, we can leverage this distinction to develop a runtime robust detector.
For example, uncertainty values are used as features to build a binary classifier as a detector.
Feinman et al. \cite{feinman2017detecting} introduced the Bayesian Uncertainty metric, employing Monte Carlo dropout to estimate uncertainty, primarily detecting adversarial examples situated near the class boundaries, while Smith et al. \cite{smith2018understanding} utilised a mutual information approach for the same purpose.
Furthermore, Hendrycks et al. \cite{hendrycks2016baseline} demonstrated that softmax prediction probabilities can be used to identify adversarial examples. They appended a decoder to reconstruct clean inputs from the softmax and jointly trained it with the baseline classifier. 
Following the hypothesis that diverse models exhibit different mistakes when confronted with the same attack inputs, Monteiro et al. \cite{monteiro2019generalizable} proposed a bimodel mismatch detection. 
Moreover, Feinman et al. \cite{feinman2017detecting} introduced kernel density estimation for each class within the training data and subsequently trained a binary classifier as a detector, utilising the density and uncertainty features associated with clean, noisy, and adversarial examples.
Although there are also few runtime monitoring methods for detecting adversarial examples in LLMs, we believe these current techniques can be extended to LLMs once we can develop an LLM detection model.
}

\subsection{Monitoring Output Failures}
As we mentioned in previous sections, although LLMs have shown strong performance in many domains~\cite{bang2023multitask, liu2023summary,jiao2023ChatGPT,sobania2023analysis, zhong2023can}, they are also found to be prone to various types of failures after scrutiny and evaluation~\cite{borji2023categorical, shen2023ChatGPT, zhao2023can}, such as factual errors~\cite{zhao2023can}, coding~\cite{liu2023your, khoury2023secure}, math~\cite{frieder2023mathematical}, and reasoning~\cite{liu2023evaluating}.
These failures can spell fatal disaster for downstream tasks, especially in safety-critical applications.
To address these issues, one way is to devise a mechanism to generate constrained outputs ~\cite{hu2017toward, kumar2020syntax, madaan2021generate}.
However, LLMs generate output by selecting appropriate words from a vocabulary rather than grabbing corresponding snippets from sources of truth, or reasoning on them.
This generative nature makes it challenging to control the output, and even more challenging to ensure that the generated output is, in fact, consistent with the information source.
Another way is to monitor the output of the models and take necessary actions.
In the following, we first summarise the limited amount of existing work on runtime monitoring of such failures and then discuss how to proceed from a future perspective.

In addition to the generative nature of LLMs, the diversity of downstream tasks also makes it extremely difficult, if not impossible, to have a general monitoring framework for such generative outputs.
Such output failures need to be addressed in a targeted manner, according to different application scenarios and the specific scientific knowledge accumulated by humans in various fields such as science and technology.
Regarding factual errors, \cite{thorne2018fever} proposed a testbed for fact verification.
However, this remains an unsolved challenge.
Similar to fact-checking, we argue that for code generation failures, the fruitful methods, techniques, and tools accumulated in the field of formal methods related to \textit{compilers design} ~\cite{aho2020compilers} and \textit{program verification}~\cite{vardi1986automata} can be adapted to check whether the generated code is executable or satisfies some specified invariants ~\cite {manna2012temporal, bensalem1996powerful, bensalem1998invest}, respectively.
As for math-related failures, existing tools in \textit{automated theorem proving}~\cite{fitting2012first, bibel2013automated} (e.g., Z3~\cite{de2008z3} and Prover9~\cite{mccune2005prover9}) may help.
\blue{
If an LLM is employed within safety-critical systems and its outputs are required to adhere to specified system safety properties, then a combination of traditional runtime monitoring and enforcement techniques~\cite{bauer2011runtime,bartocci2018lectures}, along with those~\cite{garcia2015comprehensive,alshiekh2018safe,jansen2018shielded,jansen2020safe,gu2022review} specifically developed for safe reinforcement learning, can be put into action.
This allows to detect in real-time whether the model's outputs violate predefined behavioural specifications and enforce corrective actions on the model's outputs to ensure the safe operation of the system.
}

\blue{
In order to conduct runtime monitoring for the aforementioned output errors, substantial offline or online overheads are incurred.
This is due to the requirement of establishing an auxiliary system aimed at efficiently detecting a range of output anomalies in the model.
}

Finally, we point out that the current research on the output failures of large-scale language models is still blank.
More research is needed, such as configuring a runtime monitor for the output of a specific application, or combining symbolic reasoning and causal reasoning with the model's learning process to ensure that the output avoids failures from the source.

\subsection{Perspective}
Since LLMs are still in their infancy and have many known vulnerabilities, monitoring these models in real time is a longstanding challenge.
In this section, we outline topics for future work to call on more researchers to address this challenge from three perspectives: why, what, and how.

\textit{Why does a model need to be monitored}?
The first thing we want to highlight is whether at some point the LLMs can be trained intelligent enough, so that there is no need to design a separate runtime monitor for these models.
For instance, the model is endowed with abilities to automatically detect ``illegal inputs'' (e.g., out-of-distribution inputs) and guarantee the correctness of its outputs.
From our authors' perspective,  achieving such level of intelligent models in the foreseeable future is very difficult, if not impossible.
The main reasons are as follows.
Existing LLMs are still learned from observations, i.e., a training dataset containing partial information.
There is no evidence that current learning methods can infer from parts to wholes, even in the case of massive data, nor is there evidence that a training dateset captures all the information.
Furthermore, existing learning methods do not characterize their generalization bounds but instead measure the so-called generalization error, which prevents the identification of ``illegal inputs''.
Therefore, it is necessary to monitor the model in real-time.

\textit{What should be monitored}? 
One needs to overcome various vulnerabilities listed in Section~\ref{sec:vulnerabilities} to reliably use LLMs in safety-critical applications.
Equipping the model with a corresponding runtime monitor provides a possible solution complementary to offline verification methods.
For example, there have been some works on monitoring whether the model's prediction is made for out-of-distribution inputs and whether the model's output is consistent with some existing fact base.
However, to our knowledge, there is no monitoring work on other output failures, e.g., reasoning and code errors; on intended attacks, e.g., robustness, backdoor, and data poisoning. 
Thus, we call on researchers and practitioners to investigate more in these topics.

\textit{How to better design a monitor for a model}? The state-of-the-art methods are based on the uncertainty model's predictions.
Unfortunately, low uncertainty cannot assure the model's prediction is reliable, and vice versa.
To better design monitors for LLMs, we need the following efforts.
First, some fundamental intrinsic issues of deep learning models must be better addressed, such as model implicit generalisation and decision boundaries and explainability of model decisions, which may provide more rigorous and formal characterisation and specification for building monitors.
Specific to LLMs, some special issues need to be tackled, such as the unavailability of training datasets, the non-transparency of models, the generative nature of multi-modality, etc.
Regarding specific tasks, such as the most studied problem of monitoring out-of-distribution inputs, principled methods for system design and evaluation of monitors still needs to be included, as current work is based on calibration of predictive confidence scores and evaluation on one-sided test datasets.
Last, we call for great attention to unexplored topics, such as how to monitor other trustworthiness and responsibility issues,  attacks, and unintended bugs, along with the model's social and ethical alignments with human society. 

\section{Regulations and Ethical Use}\label{sec:ethical}

\gls{vnv} provides a set of technical means to support the alignment of \gls{llm}s with human interests. However, it has been argued that constructing \gls{llm}s that cannot be abused can be impossible. This suggests that technical means are necessary, but can be insufficient. To this end, it is needed to have \emph{ethical means}, to supplement the technical means, in ensuring that the \emph{complete alignment} of the use of \gls{llm}s with human interests. In the following, we discuss several recent signs of progress. 
 
\subsection{Regulate or Ban?}

A recent debate on ``a 6-month suspension on the development \cite{sixmonthpause} vs. a regulated development'' has shown the anxiety of, and the difference of opinions from, the community upon the possibilities of  AI development being misaligned with human interests. More radical actions have also been taken. For example, Italy has reportedly banned the ChatGPT \cite{italybanChatGPT}. In a US Senate Hearing on May 2023, OpenAI CEO Sam Altman asked the government to regulate AI \cite{AltmanTestimony}. Actually, on AI regulations, major players such as the EU, US, UK, and China all have their respective approaches and initiatives, e.g., the EU's GDPR \cite{EUGDPR}, AI Act \cite{EUAIAct}. Data Act \cite{EUDataAct}, the UK's Data Protection Act \cite{UKGDPR} and pro-innovative approach to regulate AI \cite{UKProinnovative}, the US's Blueprint for an AI Bill of Rights \cite{USwhitehousebill} and AI Risk Management Framework \cite{USriskframework}, and China's regulations for recommendation algorithms \cite{ChinaRecommendationRegulation}, deep synthesis \cite{ChinaDeepSyntehsisRegulation}, and algorithm registry \cite{ChinaAlgorthimRegistry}. It is unclear (1) whether these regulations on the more general AI/ML, or other AI/ML algorithms, can automatically work for \gls{llm}s without any changes, and (2) how the regulations can be projected onto each other in a rigorous, yet operational, way. More importantly, even for general AI/ML, it  still needs to be clarified how to \emph{sufficiently and effectively} address regulatory requirements (such as robustness and transparency) with technical means. The \gls{vnv} framework proposed in this survey will be one viable solution. 

Nevertheless, significant issues raised by the \gls{llm}s, notably the ChatGPT, include copyright and privacy. The ChatGPT developers  reportedly use data from the internet, and it is unclear if the \emph{copyrights of the training data} have been carefully dealt with, especially when the ChatGPT is eventually for commercial use. Moreover, as a conversational AI, the \emph{privacy of the end users}, when engaged in a dialogue, is a serious concern. The end-users should be informed on whether and how their dialogues will be stored,  used, and redistributed. 


\subsection{Responsible AI Principles}

Responsible and accountable AI has been a topic of discussion for the past years (see e.g., \cite{EUEthicsAI,MSAIprinciples,TuringAIPrinciples}), with a gradual convergence to include properties such as transparency, explainability, fairness, robustness, security, privacy, etc. A governance framework is called for to ensure that these properties are implemented, evaluated, and monitored. A comprehensive discussion and comparison is out of the scope of this survey, but we note that, many properties are required, consistent definitions to many of them are still missing, and properties can be conflicting (i.e., the improvement of one property may compromise others). It is therefore not surprising that it can still be a long way to turn the principles into operational rules. 

Specific to \gls{llm}s, ChatGPT and the like have led to severe concerns on e.g.,  potential misuse,  unintended bias, and fair access. 
To this end, on the enterprise level, ethical principles are needed to guide the development and use of \gls{llm}s, including questioning whether something should be done rather than whether it can be done, as requested in \cite{naturecomments}. \blue{Moreover, systematic research is also called for to understand the certain to which the misuse of LLMs can lead to bad consequence, as is done in \cite{10188649} on attackers generating malware with LLMs or in \cite{sandoval2023lost} which discusses the security implication of using LLMs in generating codes.}  

\subsection{Educational Challenges}

\blue{
Verification and validation of safe and trustworthy AI models are not central to the computer science curriculum 
or data science curricula. When validation and verification of models are taught at all, it is often part of an 
AI course that emphasizes tinkering with testing data-set rather than as a systematic and rigorous discipline with a 
solid scientific foundation. We need a curriculum beyond traditional education, covering formal verification, statistics, and XAI.}

\blue{
This need for adequately trained engineers impacts industrial practice, creating inefficiencies and difficulties 
in building AI systems with safety guarantees. Engineers untrained in safety and trustworthy AI models are 
often asked to make AI models for AI-critical applications.}

\blue{
The need for a shared cultural background between AI and rigorous design communities results in fragmented 
research. They use different terminologies. For example, ``trustworthiness'' does not have the same meaning 
across communities. Conferences are separate, no interaction between the two communities. 
The educational system will take time to adapt to evolving industrial and cultural needs. 
At the least, we suggest introducing AI students to the rigorous and systematic analysis of safety and trust 
and the corresponding approaches to the design of AI-critical applications. Another short-term objective 
should be to define and promote a reference curriculum in computer science with an optional program 
for designing safe and trusted AI applications.
}

\subsection{Transparency and Explainability}

First, OpenAI's decision to not open-source GPT-3 and beyond has already led to concerns on the transparent development of AI. However, OpenAI said it plans to make more technical details available to other third parties  for them to advise on how to weigh the competitive and safety considerations against the scientific value of further transparency. \blue{Nevertheless, we have seen a trend of open-sourcing \gls{llm}s, with notably Meta's Llama 2 \cite{touvron2023llama}.}  It is also important to note that, no technical details are available on how the guardrail is designed and implemented. It will also be interesting to discuss on whether the guardrail itself should undergo a verification process.  

Second, it has been hard to interpret and explain the decisions of the deep learning models such as image classifiers. The situation becomes worsens when dealing with \gls{llm}s \cite{10.1145/3546954}, which have emergent and hard-to-explain behaviours. For example, it has been observed that adding an incarnation, such as ``Let's think step by step'', to the prompt can achieve improved responses from GPT-3. Techniques are needed to explain such a phenomenon. This calls for  extending explainable AI techniques to work with \gls{llm}s. In particular, it is necessary to consider the explanations' robustness to explain why such incarnation can lead to improved, yet different, answers. To this end, some prior works on image classifiers, such as \cite{pmlr-v161-zhao21a,huang2022safari},  can be considered. 

\section{Discussions}

\blue{
The safety and trustworthiness issues become more important with the wider adoption of machine learning, especially for \gls{llm}s, with which a large number of end users have direct interactions. Research has been significantly lagged behind, partly due to the fact that some issues become more significant for \gls{llm}s than they are for the usual machine learning models. The following are an incomplete list of research directions that we believe  require significant investments in the near future.  
\begin{itemize}
    \item Data privacy. For usual machine learning models, their training data are obtained beforehand, with many of them being made available to the public. Notable examples include the ImageNet dataset. That is, the privacy and copyright issues for the training data were not as serious. However,  \gls{llm}s' training data come directly from the internet, many of which are private information and do not have the authorisations from the data owners. On top of this, various techniques, such as prompt injection and privacy attacks, are available to leak the information. It requires a multi-disciplinary approach to deal with the data privacy issue. 
    \item Safety and trustworthiness implications. Currently, research is focused on tricking the \gls{llm}s to generate unexpected outcomes. There needs to be systematic approaches to study and measure the certain to which such unexpected outcomes might lead to bad consequences. This requires the modelling of the environment (e.g., an organisation) in which the \gls{llm}s are used, including how they are used and the consequences of all possible outcomes. A systematic study will enable the understanding of which aspects of alignments are needed and how to fine-tune the \gls{llm}s to different applications domains. 
    \item Rigorous engineering. The \gls{llm}s, in its current development, are mostly relying on the massive training data and the exceptional computational power owned by the large tech giants. Its performance currently is measured with various small scale benchmark datasets that are designed for the domain specific aspects of the abilities, for example, the mathematics, the reasoning, and so on. A rigorous engineering approach, by considering the entire development cycle including the evaluation, is needed to support the shifting of the development from extensive mode to intensive mode, and for the benefit of providing assurance cases for the applications of \gls{llm}s to safety critical domains. 
    \item Verification with provable guarantees. Empirical evaluation provides certain evidence about the performance,  but cannot be regarded as a rigorous justification, especially in safety critical domains. A mathematically well-founded proof about the performance, e.g., in the form of statistical guarantees such as chain constraint \cite{bensalem2023what}, can be useful for improving the confidence of the users. 
    \item Regulations and standards. The necessity of regulations has been commonly agreed. However, the regulations do not provide workable measures that are usually recommended in industrial standards. Compliance with regulations and standards is an important part of an assurance case to justify the safety of a product. It is  urgent for the community to come up with standards so as to release the full potential of \gls{llm}s and AI in general. 
\end{itemize}
}

\section{Conclusions}\label{sec:concl}

This paper provides an overview of the known vulnerabilities of \gls{llm}s, and discusses how the \gls{vnv} techniques might be adapted to work with them. Given the \gls{llm}s are quickly adopted by applications that have direct or indirect interactions with end users, it is imperative that the deployed \gls{llm}s undergoes sufficient verdict processes to avoid any undesirable safety and trustworthy consequences. 
\blue{Novel \gls{vnv} techniques are called for, to deal with the special characteristics of the \gls{llm}s such as the nondeterministic behaviours, the model sizes that are significantly larger than the usual machine learning models, the training dataset that is obtained from internet rather than through a careful collection process, etc. }
Multi-disciplinary development is needed to make sure that all trustworthy issues are fully considered and tackled. 

\section*{Acknowledgements}
 This project has received funding from the European Union’s Horizon 2020 research and innovation programme under grant agreement No 956123.  It is also financially supported by the U.K. EPSRC through End-to-End Conceptual Guarding of Neural Architectures [EP/T026995/1].


\bibliographystyle{abbrv}
\bibliography{sample-base}


\end{document}